\documentclass[conference,status=final]{IEEEtran}
\IEEEoverridecommandlockouts
\usepackage[svgnames,table]{xcolor}
\usepackage{graphicx}
\definecolor{lightgray}{gray}{0.9}
\definecolor{LightBlue}{rgb}{0.68, 0.85, 0.9}
\definecolor{DarkSeaGreen}{cmyk}{0.69,0,0.50,0.5}
\definecolor{DarkGreen}{rgb}{0,.5,0}
\definecolor{MyGray}{gray}{.5}
\definecolor{ForestGreen}{rgb}{0,0.71,0.04001}
\usepackage[ngerman,english]{babel}
\usepackage{xspace}
\usepackage[babel,german=quotes]{csquotes}
\usepackage{amsmath}
\usepackage{paralist}
\usepackage[shortlabels]{enumitem}
\usepackage{array}
\usepackage{subfig}
\usepackage{amssymb,amsthm}
\usepackage{booktabs}
\usepackage{tabularx}
\usepackage{multirow}
\usepackage{multicol}
\usepackage{mdframed}

\usepackage{acro}
\acsetup{
format-include-endings = true,
group-citation = true}
\DeclareAcronym{AAIP}{short = AAIP, long = Assuring Autonomy International Programme}
\DeclareAcronym{ASC}{short = ASC, long = automatic safety controller}
\DeclareAcronym{CE}{short = CE, long = critical event}
\DeclareAcronym{DTMC}{short = DTMC, long = discrete-time Markov chain}
\DeclareAcronym{FTA}{short = FTA, long = fault tree analysis}
\DeclareAcronym{FMEA}{short = FMEA, long = failure mode effects analysis}
\DeclareAcronym{HazOp}{short = HazOp, long = hazard operability studies}
\DeclareAcronym{HRC}{short = HRC, long = human-robot collaboration}
\DeclareAcronym{IRS}{short = IRS, long = industrial robot system}
\DeclareAcronym{LTS}{short = LTS, long = labelled transition system}
\DeclareAcronym{MC}{short = MC, long = Markov chain}
\DeclareAcronym{MDP}{short = MDP, long = Markov decision process}
\DeclareAcronym{PCTL}{short = PCTL, long = probabilistic computation tree logic}
\DeclareAcronym{pCTMC}{short = pCTMC, long = parametric continuous-time Markov chain}
\DeclareAcronym{pGCL}{short = pGCL, long = probabilistic guarded command language}
\DeclareAcronym{pMDP}{short = pMDP, long = parametric Markov decision process}
\DeclareAcronym{POMDP}{short = POMDP, long = partially observable Markov decision process}
\DeclareAcronym{PTA}{short = PTA, long = probabilistic timed automaton}
\DeclareAcronym{pPTA}{short = pPTA, long = parametric probabilistic timed automaton}
\DeclareAcronym{UoY}{short = UoY, long = University of York}

\usepackage[colorlinks,allcolors=Blue]{hyperref}
\usepackage{cleveref}
\Crefname{section}{Sec.}{Secs.}
\Crefname{table}{Tab.}{Tabs.}
\Crefname{figure}{Fig.}{Figs.}
\Crefname{equation}{Eq.}{Eqs.}
\Crefname{claim}{Claim}{Claims}
\Crefname{paragraph}{Section}{Sections}
\Crefname{theorem}{Theorem}{Theorems}
\Crefname{lemma}{Lemma}{Lemmas}
\Crefname{equation}{Formula}{Formulas}
\Crefname{proof}{Proof}{Proofs}
\Crefname{property}{Property}{Properties}
\Crefname{lstlisting}{Listing}{Listings}

\usepackage{tikz}
\usetikzlibrary{babel, arrows, arrows.meta, positioning, automata, shapes,
  calc, decorations.pathreplacing, decorations.pathmorphing, graphs}

\newcommand{\mgitem}[1][1]{\tikz[baseline=(#1.base),every
  node/.style={transform shape}, scale=.9] \node[stlab] (#1) {#1};}

\usepackage{ctable}

\newcommand{\event}[1][]{\mathit{#1}}
\newcommand{\rfct}[1][f]{\mathsf{#1}}
\newcommand{\rapac}[2][]{\event[#2]_{#1}}
\newcommand{\acend}[2][]{\rapac[#1]{e}^{\rfct[#2]}}

\newcommand{\acmit}[2][]{\rapac[#1]{m}^{\rfct[#2]}}

\newcommand{\inact}[1][\rfct]{0^{#1}}
\newcommand{\activ}[1][f]{\mathit{#1}}
\newcommand{\mitig}[1][\rfct]{\overline{#1}}
\newcommand{\mishp}[1][\rfct]{\underline{#1}}

\newcommand{\yapkeyw}[1]{\textcolor{blue}{\textbf{#1}}}

\newtheorem{definition}{Definition}
\newmdtheoremenv[%
hidealllines=true,
leftmargin=-4pt,rightmargin=-4pt,
backgroundcolor=gray!20,%
innertopmargin=0pt,%
innerbottommargin=4pt,%
innerleftmargin=4pt,%
innerrightmargin=4pt,%
font={\small},
ntheorem]{example}{Example}

\usepackage{cite}
\usepackage[font={small}]{caption}
\def\BibTeX{{\rm B\kern-.05em{\sc i\kern-.025em b}\kern-.08em
    T\kern-.1667em\lower.7ex\hbox{E}\kern-.125emX}}
\usepackage{flushend}

\tikzset{
  stlab/.style={circle,fill=cyan!30},
}

\begin{document}
\pagestyle{plain}
\thispagestyle{plain}
\title{Safety Controller Synthesis for Collaborative Robots}
\author{\IEEEauthorblockN{
    Mario Gleirscher\IEEEauthorrefmark{1}\IEEEauthorrefmark{2},
    Radu Calinescu\IEEEauthorrefmark{1}\IEEEauthorrefmark{2}}
  \IEEEauthorblockA{\IEEEauthorrefmark{1}\textit{Assuring Autonomy
      International Programme, University of York}, York, UK}
  \IEEEauthorblockA{\IEEEauthorrefmark{2}\textit{Department of
      Computer Science, University of York}, York, UK\\ 
    mario.gleirscher,radu.calinescu@york.ac.uk}
}

\maketitle

\begin{abstract}
  In \ac{HRC}, software-based \acp{ASC} are used 
  in various
  forms~(e.g.\xspace shutdown mechanisms, emergency brakes, interlocks) to
  improve operational safety.  Complex robotic tasks and increasingly
  close human-robot interaction pose new challenges to \ac{ASC} developers and
  certification authorities. Key among these challenges is the need 
  to assure the correctness 
  of \acp{ASC}
  under reasonably weak
  assumptions. %
  To address this need, we introduce and evaluate a tool-supported \ac{ASC} synthesis
  method for \ac{HRC} in manufacturing. Our \ac{ASC} synthesis is:
  \begin{inparaenum}[(i)]
  \item informed by the manufacturing process, risk
    analysis, and regulations; 
  \item formally verified against correctness criteria; and
  \item selected from a design space of feasible controllers according to a set of 
    optimality criteria.
  \end{inparaenum}
  The synthesised \ac{ASC} can detect the occurrence of hazards, move
  the process into a safe state, and, in certain 
  circumstances, %
  return the process to an operational state 
  from which it can resume its original task.
\end{abstract}
\begin{IEEEkeywords}
  Controller synthesis, human-robot collaboration, software
  engineering, probabilistic model checking.
\end{IEEEkeywords}

\section{Introduction} 
\label{sec:intro}

An effective collaboration between %
\acp{IRS} and
humans~\cite{Nicolaisen1985-OccupationalSafetyIndustrial,
  Jones1986-StudySafetyProduction} %
can leverage 
their complementary skills, %
but is difficult to achieve 
because of uncontrolled \emph{hazards} and unexploited sensing,
tracking, and safety
measures~\cite{Santis2008-atlasphysicalhumanrobot}.
Such hazards have been studied since the 1970s, resulting in elaborate
risk taxonomies based on workspaces, tasks, and human body
regions~\cite{Sugimoto1977-SafetyEngineeringIndustrial,
  Jones1986-StudySafetyProduction,
  Alami2006-Safedependablephysical,
  Haddadin2009-RequirementsSafeRobots,
  Wang2017-Humanrobotcollaborativeassembly, %
  Kaiser2018-SafetyRelatedRisks,
  Matthias2011-Safetycollaborativeindustrial,
  Marvel2015-CharacterizingTaskBased}.  The majority are \emph{impact
  hazards}~(e.g.\xspace unexpected movement, reach beyond area, dangerous
workpieces, hazardous manipulation), \emph{trapping hazards}~(e.g.\xspace
operator in cage), and \emph{failing equipment}.

Addressing these hazards involves 
the examination of each \emph{mode of operation}~(e.g.\xspace normal,
maintenance) for its hazardous behaviour, and the use of automatic safety controllers (\acp{ASC})
to trigger mode-specific \emph{safety measures}~\cite{Jones1986-StudySafetyProduction}. %
Malfunction diagnostics~(e.g.\xspace fault detection, wear-out monitoring)  %
can further inform the \ac{ASC}.  
As shown in \Cref{tab:measures}, a 
variety of safety 
measures~\cite{Santis2008-atlasphysicalhumanrobot} %
can \emph{prevent or mitigate} hazards and
accidents by reducing the \emph{probability of their occurrence} and
the \emph{severity of their consequences}.  There are
\emph{functional}
measures using electronic equipment~(e.g.\xspace speed \& separation
monitoring) and \emph{intrinsic} measures not using such
equipment~(e.g.\xspace fence, flexible surface).  Functional measures
focusing on the correctness and reliability of a controller
are called \emph{dependability}
measures~\cite{Alami2006-Safedependablephysical,
  Avizienis2004-Basicconceptstaxonomy}.  Functional measures are said
to be \emph{passive} if they focus on severity reduction~(e.g.\xspace
force-feedback control), %
\emph{active} otherwise~(e.g.\xspace safety-rated monitored
stop). %

\begin{table}[b]
  \centering
  \caption{\acs{IRS} safety measures by stage of causal chain}
  \label{tab:measures}
  \footnotesize
  \begin{tabularx}{\columnwidth}{%
    >{\bfseries\hsize=.15\hsize}X
    >{\hsize=.31\hsize}X
    >{\hsize=.55\hsize}X}
  \specialrule{\heavyrulewidth}{\aboverulesep}{0em}
  \rowcolor{lightgray}{\large$\phantom{\mid}$\hspace{-.25em}}
  \textbf{Stage}
  & \textbf{Type of Measure}
  & \textbf{Examples}
  \\\specialrule{\lightrulewidth}{0em}{\belowrulesep}
  \multirow{2}{2cm}{Hazard prevention}
  & 1. safeguard/barrier
  & fence, interlock
  \\
  & 2. IT safety %
  & \emph{verified safety controller}
  \\
  & 3. IT security %
  & security-verified (safety) controller
  \\\midrule
  \multirow{2}{1.5cm}{Hazard mitigation \&
    accident prevention}
  & 4. reliability 
  & fault-tolerant scene interpretation 
  \\
  & 5. workspace intrusion detection
  & speed \& separation monitoring,
  safety-rated monitored stop
  \\
  & 6. shift of control
  & hand-guided operation
  \\\midrule
  \multirow{2}{1.5cm}{Accident mitigation (alleviation)}
  & 7. power \& force limitation
  & low weight parts, flexible surfaces; variable impedance,
  touch-sensitive, \& force-feedback control 
  \\
  & 8. system halt
  & emergency stop, dead-man's switch
  \\\bottomrule
  \end{tabularx}
\end{table}

The standardisation of safety requirements for
\acp{IRS}~\cite{Sugimoto1977-SafetyEngineeringIndustrial} culminated
in ANSI/RIA R15.06, ISO~10218~\cite{ISO10218}, 13482, and 15066.
According to ISO~10218, an \ac{IRS} comprises a \emph{robot arm}, a
\emph{robot controller}, an \emph{end-effector}, and a
\emph{workpiece}~(see, e.g.\xspace\Cref{fig:setting-conceptual} below).  In
\emph{collaborative operation}, the operator and the \ac{IRS} (called
a \emph{cobot}\cite{Gillespie2001-generalframeworkcobot}) can occupy
the \emph{collaborative workspace}~(i.e.,\xspace a subset of the
\emph{safeguarded workspace}) simultaneously while the \ac{IRS} is
performing \emph{tasks}~\cite{ISO15066}. %
Based on that, ISO~15066 recommends four \emph{safety modes},
described and combined with work layouts in
\cite{Kaiser2018-SafetyRelatedRisks,
  Villani2018-Surveyhumanrobotcollaboration}:
\begin{itemize}
\item \emph{safety-rated monitored stop}~(powered but no simultaneous
  activity of robot and operator in shared workspace),
\item \emph{hand-guided operation}~(zero-gravity
  control, guided by an operator, no actuation 
  without operator input),
\item \emph{speed \& separation monitoring}~(speed continuously
  adapted to distance of robot and operator), and
\item \emph{power \& force limiting} (reduced impact on the
  human body, a robot's power and applied forces are limited).
\end{itemize}
In the following, we highlight recent challenges and explain how our
work addresses these.

\subsubsection*{Challenges}
\label{sec:safety-req}

Since the 1980s, tele-programming and simulation have led to a
reduction of hazard exposure.  However, guarding arrangements
interfere with manufacturing processes and mobile robots.  Complex
tasks require continuous and close human-robot interaction~(e.g.\xspace mutual take-over of
tasks), mutual clarification of intent, and 
trading off risk~\cite{Hayes2013-ChallengesSharedEnvironment,
  Villani2018-Surveyhumanrobotcollaboration}.
Robot movements need to
be predictable and impacts on the human body need to be
attenuated~(e.g.\xspace speed \& separation monitoring requires stereo vision
and laser scanners to distinguish safety zones).  Engineers need to
consider a variety of complex failure modes.  This situation implies
\emph{requirements and design spaces} for \acp{ASC}, so engineers want
to answer questions such as:
\begin{itemize}
\item %
  Which \ac{ASC} design minimises the probability of incidents in
  presence of human and sensor errors?
\item %
  Which design minimises nuisance to the human, maximises productivity, etc.\ %
  while maintaining safety?
\item %
  Does a selected controller correctly handle hazards when detected
  and return the system 
  to a useful safe state?
\end{itemize}

\subsubsection*{Contributions}

We introduce a tool-supported method for the synthesis of discrete-event
\acp{ASC} that meet safety requirements and optimise process
performance for human-robot \emph{cooperation} (alternative use of shared workspace) 
and \emph{collaboration}
   (simultaneous use of shared workspace, with close interaction)
\cite{Kaiser2018-SafetyRelatedRisks,
  Helms2002-robatworkRobotassistant}. %
We model the manufacturing process and its safety analysis as a
\ac{MDP} and %
select a correct-by-construction \ac{ASC} from a set of \ac{MDP}
policies.  We extend our notion of \emph{risk
  structures}~\cite{Gleirscher2017-FVAV} and our tool
\textsc{Yap}\xspace~\cite{Gleirscher-YapManual}. This simplifies the modelling
of activities and actors, \aclp{CE}~(\acsp{CE}, e.g.\xspace hazards),
mitigations~(e.g.\xspace safety mode changes) and reward structures for risk
optimisation; and automates the translation of risk
structures into \acp{MDP}.  %
Our approach facilitates the verification of safety of the \ac{MDP}
and of \emph{probabilistic reach-avoid} properties of a selected
policy.  A verified \ac{ASC} detects hazards and controls their
mitigation by
\begin{inparaenum}[(i)]
\item the execution of a safety function,
\item a transition to a safer mode, or
\item a transition to a safer activity.
\end{inparaenum}

\subsubsection*{Overview}

\Cref{sec:related-work} discusses related work,
\Cref{sec:running-example} introduces our case study as a running
example, and  \Cref{sec:prelim} provides the theoretical background.
We describe and evaluate the \ac{ASC} synthesis method in \Cref{sec:approach} and \Cref{sec:eval}, respectively, and we conclude with a short summary in \Cref{sec:conclusion}. 

\section{Related Work}
\label{sec:related-work}

To the best of our knowledge, our method is the first end-to-end
approach to synthesising \acp{ASC} for handling multiple risks in
\ac{HRC} for manufacturing processes.

Askarpour~et~al.~\cite{Askarpour2016-SAFERHRCSafety} discuss a
discrete-event formalisation of a work
cell %
in the linear-time temporal language TRIO. %
Actions are specified as $\mathit{pre/inv/post}$-triples~(with a
safety $\mathit{inv}$ariant) for contract-based reasoning with the SAT
solver Zot.  In contrast, our approach builds on \ac{pGCL}, separating
action modelling from property specification.  Beyond counterexamples
for model repair, our approach yields an executable
policy. %
While their use of a priority parameter helps to abstract from
unnecessary state variables, we propose guards to implement 
flexible individual action orderings.
Moreover, violations of $\mathit{inv}$ lead to
pausing the cell whereas our approach can deal with multiple
mitigation options offering a variety of safety responses.
  
For generic robot applications,
Orlandini~et~al.~\cite{Orlandini2013-ControllerSynthesisSafety} employ
the action language PDDL for modelling and timed game automata for
controller synthesis.  The model checker UPPAAL-TIGA is used for
verifying~(i.e.,\xspace finding winning strategies for) %
reach-avoid properties of type
$\mathop{\textbf{A}}(\mathit{safe} \mathop{\textbf{U}} \mathit{goal})$.  While game
solving could enhance our verification approach, our method focuses on
guidance in risk modelling for safely optimised \ac{HRC} performance.
Cesta~et~al.~\cite{Cesta2016-Towardsplanningbased} present an approach
to synthesise controllers~(i.e.,\xspace plans) for \ac{HRC} applications using
a timeline-based PDDL planner.
While they distinguish controllable~(i.e.,\xspace duration known) from
uncontrollable actions~(i.e.,\xspace duration unknown), an important aspect of
\ac{HRC} modelling, their focus is on task planning and scheduling
rather than on risk modelling for verified synthesis of \acp{ASC}.

Heinzmann and
Zelinsky~\cite{Heinzmann2003-QuantitativeSafetyGuarantees} %
propose a power \& force limiting mode
always active during an \ac{HRC} activity described as a discrete-event
controller.  
Long~et~al.~\cite{Long2018-industrialsecuritysystem} 
propose a speed \& separation monitoring scheme with
nominal~(max.~velocity), reduced~(speed limiting), and
passive~(hand-guided operation) %
safety modes.
While these authors do not aim at synthesis or task modelling, their
elaborate safety modes may serve as a target platform to our
multi-risk synthesis approach.

\section{Running Example: Manufacturing Cobots}
\label{sec:running-example}

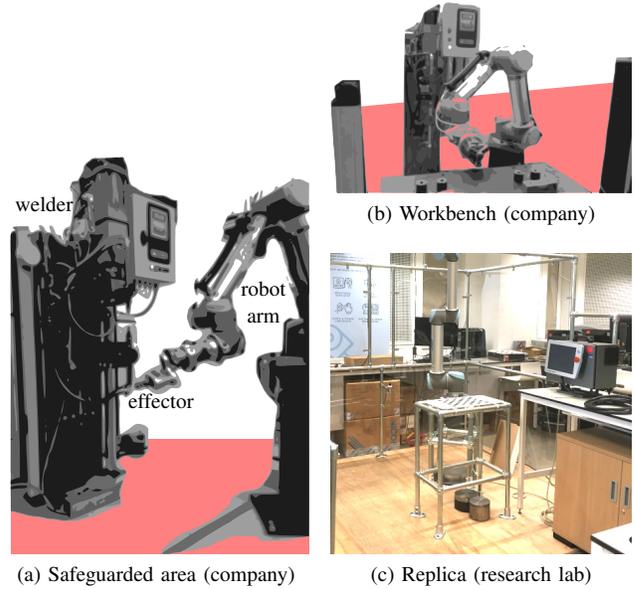
\begin{figure}
  \vspace{-2em}
  \centering
  \makeatletter
  \subfloat[Safeguarded area (company)]{
    {\footnotesize
      \def\svgwidth{3.97cm}
\begingroup%
  \makeatletter%
  \providecommand\color[2][]{%
    \renewcommand\color[2][]{}%
  }%
  \providecommand\transparent[1]{%
    \renewcommand\transparent[1]{}%
  }%
  \providecommand\rotatebox[2]{#2}%
  \newcommand*\fsize{\dimexpr\f@size pt\relax}%
  \newcommand*\lineheight[1]{\fontsize{\fsize}{#1\fsize}\selectfont}%
  \ifx\svgwidth\undefined%
    \setlength{\unitlength}{226.87545776bp}%
    \ifx\svgscale\undefined%
      \relax%
    \else%
      \setlength{\unitlength}{\unitlength * \real{\svgscale}}%
    \fi%
  \else%
    \setlength{\unitlength}{\svgwidth}%
  \fi%
  \global\let\svgwidth\undefined%
  \global\let\svgscale\undefined%
  \makeatother%
  \begin{picture}(1,1.33155289)%
    \lineheight{1}%
    \setlength\tabcolsep{0pt}%
    \put(0,0){\includegraphics[width=\unitlength,page=1]{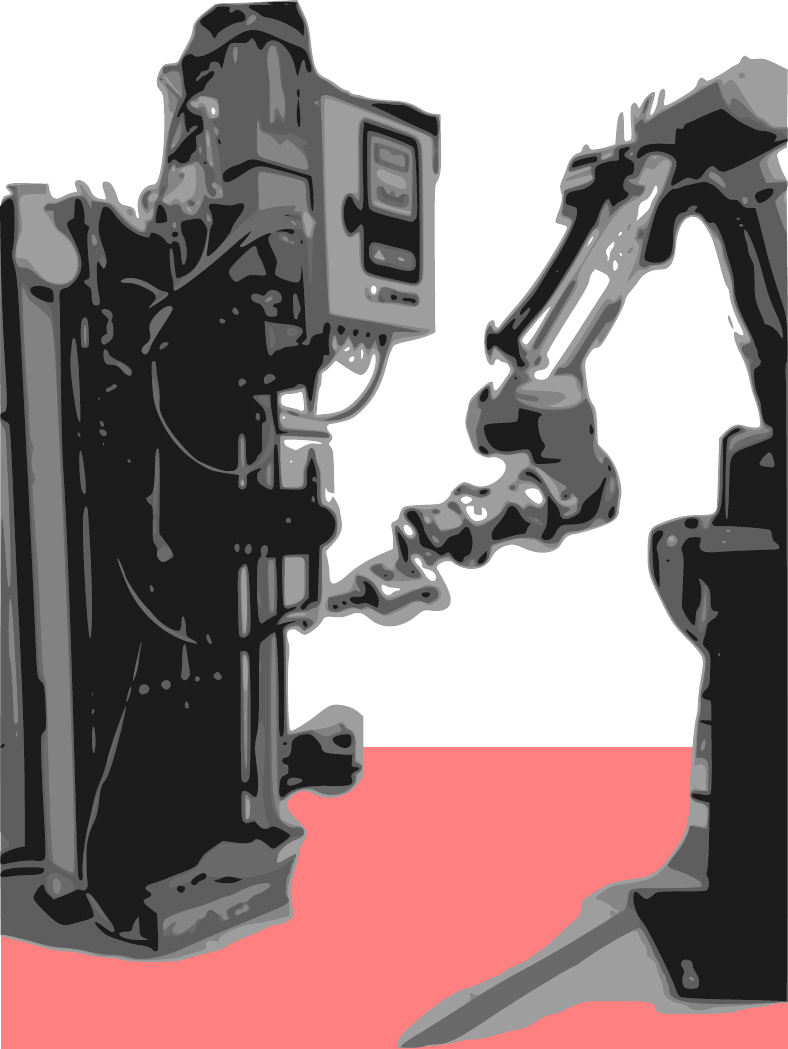}}%
    \put(0.01642296,1.15038127){\color[rgb]{0,0,0}\makebox(0,0)[lt]{\lineheight{1.25}\smash{\begin{tabular}[t]{l}welder\end{tabular}}}}%
    \put(0.8474964,0.86574726){\color[rgb]{0,0,0}\makebox(0,0)[t]{\lineheight{1.25}\smash{\begin{tabular}[t]{c}robot\\arm\end{tabular}}}}%
    \put(0.39497232,0.48872829){\color[rgb]{0,0,0}\makebox(0,0)[lt]{\lineheight{1.25}\smash{\begin{tabular}[t]{l}effector\end{tabular}}}}%
  \end{picture}%
\endgroup%
}
    \label{fig:setting-actual2}}
  \makeatother
  \hfill
  \parbox[b]{4.5cm}{
    \subfloat[Workbench (company)]{
      \includegraphics[width=4cm]{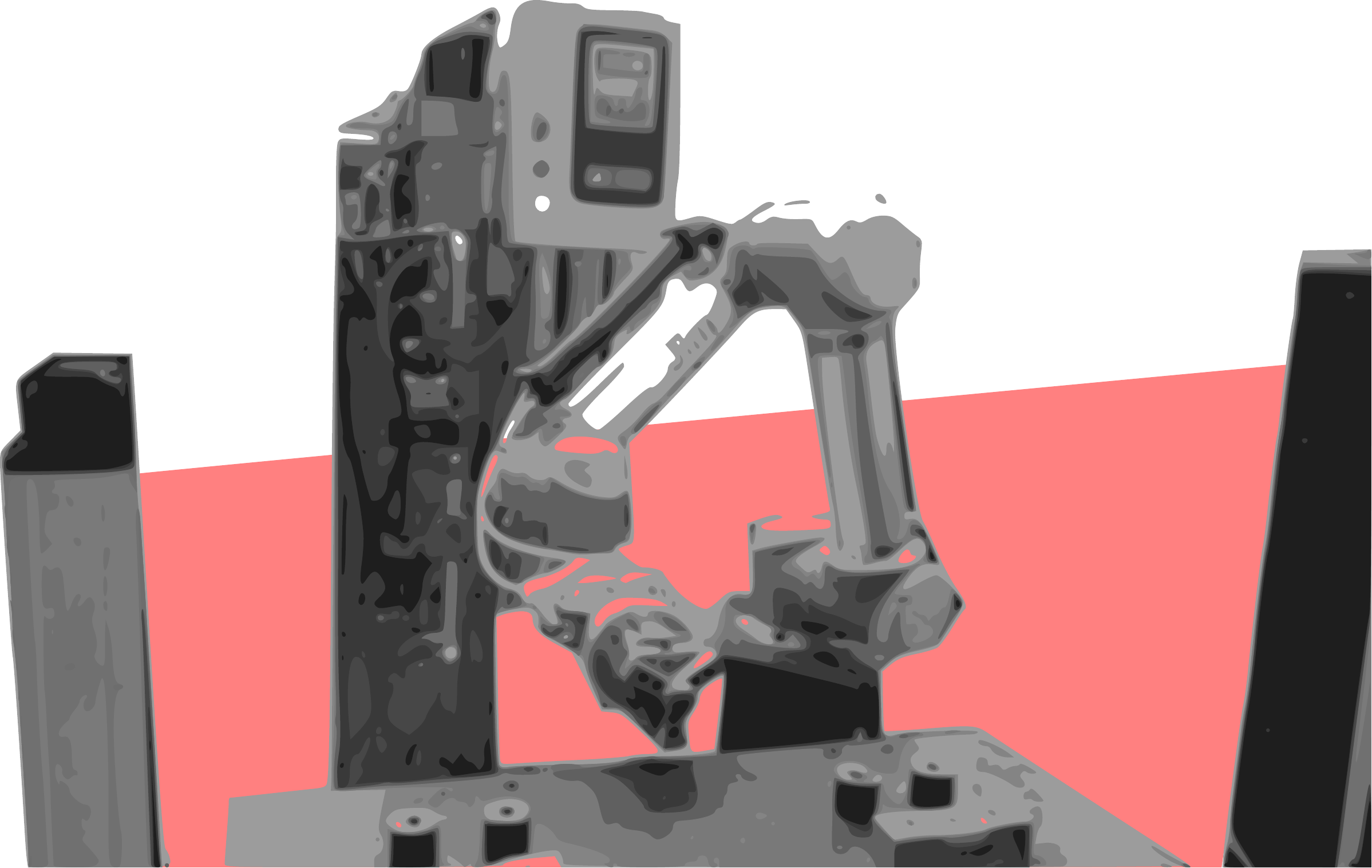}
      \label{fig:setting-actual1}}\\ 
    \subfloat[Replica (research lab)]{
      \includegraphics[width=4cm,trim=140 100 0 240,clip]{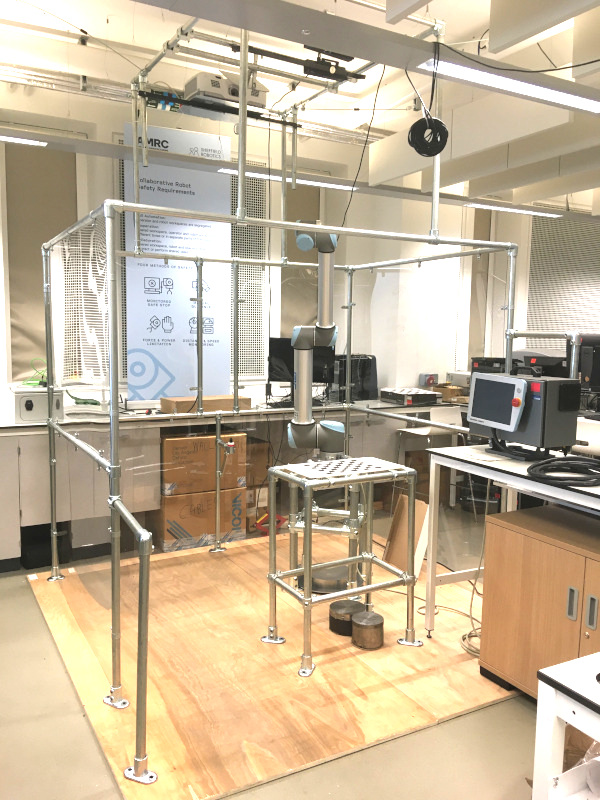}
      \label{fig:setting-replica}}}
  \caption{Actual (a, b) and replicated (c) cobot setting \label{fig:cobot-all}}
  \vspace{-1em}
\end{figure}

\begin{figure}
  \centering
  \subfloat[\ac{HRC} setting (conceptual, top view)]{%
\includegraphics[]{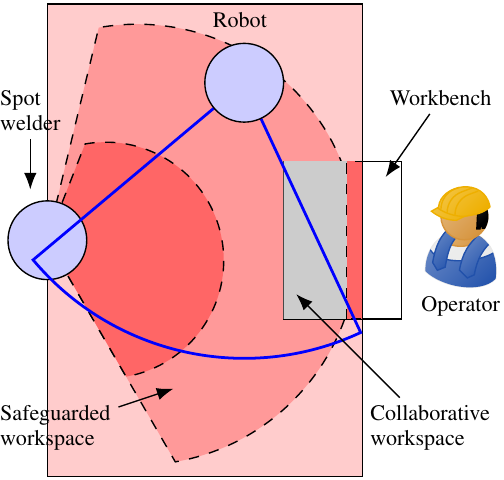}

     \label{fig:setting-conceptual}}
  \hfill
  \subfloat[Process activities]{\includegraphics[width=.14\textwidth]{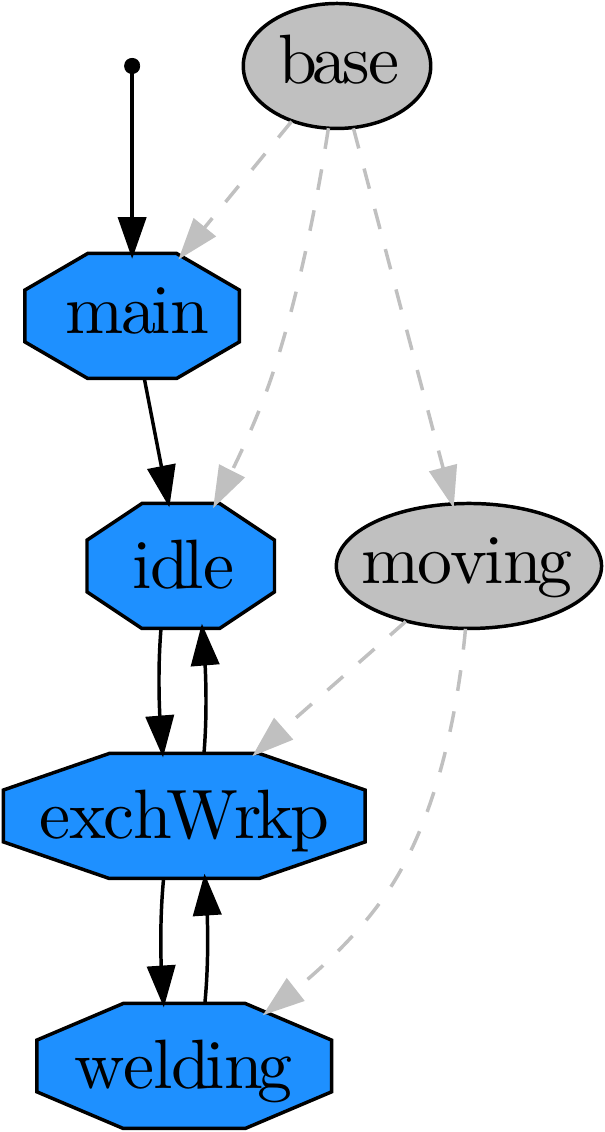}
      \label{fig:setting-activities}}
  \caption{Conceptual setting (a) and activities in the manufacturing
    process (b) performed by the operator, the robot, and the welder~(in blue),
    classified by the activity groups \emph{moving} and \emph{base}~(in gray)
    \label{fig:setting}}
  \vspace{-1em}
\end{figure}

\Cref{fig:cobot-all} 
shows an \ac{IRS} \emph{manufacturing cell} at a UK company (with the pictures anonymised for confidentiality reasons) and replicated in a testbed at the
University of Sheffield~(\Cref{fig:setting-replica}).  The corresponding
process~(call it $\mathcal{P}$) consists of
activities~(\Cref{fig:setting-activities}) collaboratively repeated by
an \emph{operator}, a stationary \emph{robotic arm}, and a
spot \emph{welder}~(\Cref{fig:setting-conceptual}).
Previous safety analysis~(i.e.,\xspace hazard identification, risk assessment,
requirements derivation) resulted in two sensors~(i.e.,\xspace a range finder
in \Cref{fig:setting-actual2} and a light barrier in
\Cref{fig:setting-actual1}, indicated in red) triggering an emergency
stop if a person approaches the welder or enters the workbench while
the robot or welder are active.
\Cref{tab:riskana} shows our partial safety analysis of the cell
following the guidance in~\Cref{sec:intro}.  The right column
specifies safety goals against each accident and controller
requirements~(e.g.\xspace mode-switch requirements) handling each latent
cause in the left column, and indicating how the hazard is to be removed.

\begin{table*}
  \centering
  \caption{Our partial safety analysis of the manufacturing cell
    referring to the measures recommended in ISO~15066}
  \label{tab:riskana}
  \footnotesize
  \begin{tabularx}{\textwidth}{l>{\hsize=.35\hsize}X>{\hsize=.65\hsize}X}
    \toprule
    \textbf{Id}
    & \textbf{Critical Event} (Risk Factor)
    & \textbf{Safety Requirement}
    \\\specialrule{\lightrulewidth}{\aboverulesep}{0em}
    \rowcolor{lightgray}{\large$\phantom{\mid}$\hspace{-.25em}}
    & \textbf{Accident} (to be \emph{prevented} or \emph{alleviated})
    & \textbf{Safety Goal}
    \\\specialrule{\lightrulewidth}{0em}{\belowrulesep}
    \emph{RC} 
    & \underline{R}obot arm harshly \underline{C}ollides with operator
    & The robot shall \emph{avoid} harsh active collisions with the
    operator.
    \\
    \emph{WS}
    & \underline{W}elding \underline{S}parks cause operator injuries %
    & The welding process shall \emph{reduce} sparks injuring the operator.
    \\
    \emph{RT}
    & \underline{R}obot arm \underline{T}ouches the operator
    & The robot shall \emph{avoid} active contact with the operator.
    \\\specialrule{\lightrulewidth}{\aboverulesep}{0em}
    \rowcolor{lightgray}{\large$\phantom{\mid}$\hspace{-.25em}}
    & \textbf{Latent Cause}~(to be \emph{mitigated} timely)$^{\dagger}$
    & \textbf{Controller Requirement}
    \\\specialrule{\lightrulewidth}{0em}{\belowrulesep}
    \emph{HRW}
    & \underline{H}uman operator and \underline{R}obot use \underline{W}orkbench at the same time
    & (m) The robot shall perform a \emph{safety-rated monitored stop}
    and (r) resume \emph{normal operation} after the \emph{operator} has left the
    \emph{shared workbench}. 
    \\
    \emph{HW}
    & \underline{H}uman operator is entering the \underline{W}orkbench %
    while the robot is away from the bench
    & (m) If the robot moves a workpiece to the bench then
    it shall switch to \emph{power \& force limiting} mode and (r) resume
    \emph{normal operation} after the \emph{operator} has left the
    \emph{workbench}.
    \\
    \emph{HS}
    & \underline{H}uman operator has entered the \underline{S}afeguarded area while robot 
    moving or welder active
    & (m) The \emph{welder} shall be \emph{switched off}, the
    \emph{robot} to \emph{speed \& separation
      monitoring}. (r) Both shall resume normal
    mode after the operator has left and acknowledged the
    notification.
    \\
    \emph{HC}
    & \underline{H}uman operator is \underline{C}lose to the welding spot while robot 
    working and welder active
    & (m) The \emph{welder} shall be \emph{switched off}, the
    \emph{robot} to \emph{safety-rated monitored stop}. (r) Both shall
    resume \emph{normal or idle mode with a reset procedure} after the
    operator has left.  
    \\\bottomrule
    \multicolumn{3}{l}{$^{\dagger}$ m\dots mitigation requirement, r\dots
      resumption requirement}
  \end{tabularx}
  \vspace{-2em}
\end{table*}

\section{Preliminaries}
\label{sec:prelim}

Our method uses \acp{MDP} as a formal model
of $\mathcal{P}$, and \ac{MDP} policies as the design space for
controller synthesis.
\begin{definition}{\acf{MDP}.}
  Given all distributions $Dist(\alpha_{\mathcal{P}})$ over an action
  alphabet $\alpha_{\mathcal{P}}$ of a process $\mathcal{P}$, an
  \ac{MDP} is a tuple
  $\mathcal{M} = (S,s_0,\alpha_{\mathcal{P}},\delta_{\mathcal{P}},L)$
  with a set $S$ of states, an initial state $s_0\in S$, a
  probabilistic transition function
  $\delta_{\mathcal{P}}\colon S\times \alpha_{\mathcal{P}} \to
  Dist(\alpha_{\mathcal{P}})$, and a map
  $L\colon S \to 2^{AP}$ labelling $S$ with atomic propositions
  $AP$~\cite{Forejt2011-AutomatedVerificationTechniques}.
\end{definition}
Given a map $A\colon S \to 2^{\alpha_{\mathcal{P}}}$, $|A(s)| > 1$
signifies non-deterministic choice in $s$.  Its resolution 
for $S$ forms a \emph{policy}.  
\begin{definition}{Memoryless Policy.}
  A \emph{memoryless policy} is a map
  $\pi\colon S \to Dist(\alpha_{\mathcal{P}})$ s.t.\xspace\
  $\pi(s)(a)>0 \Rightarrow a\in A(s)$.  $\pi$ is
  \emph{deterministic} if $\forall s\in S$ $\exists a\in A(s)\colon$
  $\pi(s)(a) = 1 \land \forall
  a'\in\alpha_{\mathcal{P}}\setminus\{a\}\colon \pi(s)(a') = 0$.
\end{definition}
The following discussion is restricted to deterministic memoryless
policies.  Let $\Pi_{\mathcal{M}}$ be the set of all such policies for
$\mathcal{M}$.  Then, \emph{action rewards} defined by a map
$r^q_{action}\colon S \times \alpha_{\mathcal{P}} \to \mathbb{R}_{\geq
  0}$ allow the assessment of $\Pi_{\mathcal{M}}$ based on a quantity
$q$.

\emph{Verification} of $\mathcal{M}$ is based on \ac{PCTL} whose
properties over $AP$ are formed by
\[
  \phi ::=
  \top
  \mid a
  \mid \neg \phi
  \mid \phi \land \phi
  \mid \mathop{\textbf{E}} \phi
  \mid \mathop{\textbf{A}} \varphi
  \quad\text{and}\quad
  \varphi ::= \mathop{\textbf{X}} \phi
  \mid \phi \mathop{\textbf{U}} \phi
\]
with $a\in AP$; an optional bound $b\in\mathbb{N}_+$ for
$\mathop{\textbf{U}}^{\sim b}$ with $\sim\;\in\{<,\leq,=,\geq\}$; the
quantification operators $\mathop{\textbf{P}}_{\sim b\mid =?}\varphi$ to verify
(or with $=?$, to quantify) probabilities,
$\mathop{\textbf{S}}_{\sim b\mid =?}[a]$ to determine long-run probabilities,
$\mathop{\textbf{R}}^q_{\sim b\mid[\min\mid
  \max]=?}[\mathop{\textbf{F}}\phi\mid\mathop{\textbf{C}}^{[\sim b]}]$ to calculate
reachability and accumulative action rewards, and the abbreviations
$\mathop{\textbf{F}}\phi \equiv \top\mathop{\textbf{U}}\phi$,
$\mathop{\textbf{G}}\phi \equiv \neg\mathop{\textbf{F}}\neg\phi$, and
$\phi\mathop{\textbf{W}}\psi\equiv \phi\mathop{\textbf{U}}\psi \lor
\mathop{\textbf{G}}\phi$.  For sake of brevity, consider the treatment of
\ac{PCTL} in \cite{Baier2008-PrinciplesModelChecking,
  Forejt2011-AutomatedVerificationTechniques}.

The concise definition of $\delta_{\mathcal{P}}$, the behaviour of
$\mathcal{P}$, is facilitated by
\textsc{PRISM}\xspace's~\cite{Forejt2011-AutomatedVerificationTechniques} \ac{pGCL}.
Guarded commands are of the form
$[\alpha]\; \gamma \longrightarrow \upsilon$ where $\alpha$ is an
event label and $\upsilon$ a probabilistic update applicable to
$s\in S$ only if $s\models\gamma$, where $\gamma$ is an expression in
the propositional fragment of \ac{PCTL}.\footnote{We use
  $\longrightarrow$ to separate guard and update expressions and
  $\rightarrow$ both for logical implication and the definition of
  mappings.}  Generally,
$\upsilon ::= \pi_1\colon \upsilon_1 + \dots + \pi_n\colon \upsilon_n$
with $\Sigma_{i\in 1..n}\pi_i = 1$ and assignments $\upsilon_i$ to
state variables of type $\mathbb{B}$, $\mathbb{N}$, or $\mathbb{R}$.

For \emph{safety analysis}, we view the cell in
\Cref{sec:running-example} as %
a \emph{process} $\mathcal{P}$, monitored and influenced by an
\ac{ASC} to mitigate \emph{hazards} and prevent \emph{accidents}.  An
\emph{accident} $a\in S$
is an undesired consequence reachable from
a set $\Xi\subset S$ forming the \emph{causes} of $a$.
The fraction of a cause $c\in \Xi$ not related to the operator
is called a \emph{hazard}
$\activ[h]$~\cite{Leveson1995-SafewareSystemSafety,
  Leveson2012-EngineeringSaferWorld}.
We call $c$ \emph{latent}\footnote{As opposed to \emph{immediate}
  causes reducing the possibilities of risk handling.} if there are
sufficient resources~(e.g.\xspace time for removing $\activ[h]$ by transition
to $s\not\in \Xi$) to prevent the accident.
$\activ[h]$ includes states in $S\setminus \Xi$ being critical because
certain events~(e.g.\xspace an operator action) cause a transition to $\Xi$,
and possibly $a$, if $\activ[h]$ stays active, further conditions
hold, and no safety measures are put in place timely.

\label{sec:yap}
\emph{Risk modelling} can be facilitated by specifying risk factors and
combining them into \emph{risk structures}~\cite{Gleirscher2017-FVAV}.
A \emph{risk factor} $\rfct$ is a \ac{LTS} modelling the life cycle of
a critical event~(i.e.,\xspace hazard, cause, mishap).  $\rfct$ has the phases
\emph{inactive}~($\inact{}$), \emph{active}~($\activ{}$), and
\emph{mitigated}~($\mitig{}$) and transitions between these phases
signifying \emph{endangerment} events~($\acend{}$) and
\emph{mitigation}~($\acmit{}$) and \emph{resumption}~($\acmit[r]{}$)
actions.  Let $F$ be a set of factors, e.g.\xspace the ones in column
\textbf{Id} in \Cref{tab:riskana}.  The Cartesian product of the
phases of the factors in $F$ yields the \emph{risk space}~$R(F)$.  To
utilise factor \acp{LTS} for the translation of \ac{ASC} designs into
\ac{pGCL}, we further develop the notion of risk factors in
\Cref{sec:appl-cobots-proj} as part of our contribution.
  
\section{Approach: Safety Controller Synthesis}
\label{sec:approach}

\begin{figure}
  \centering
\includegraphics[]{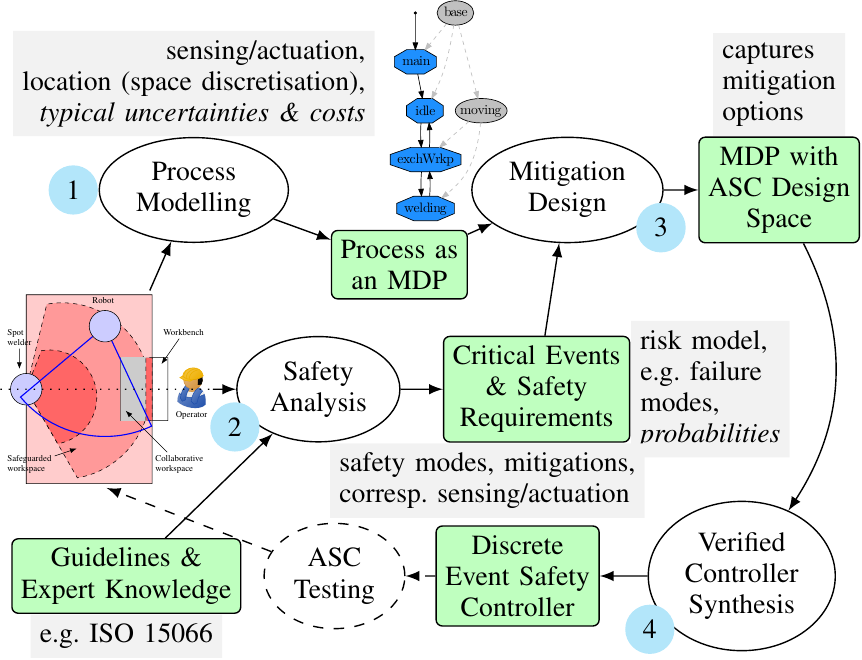}

   \caption{Main
    \protect\tikz[baseline=(X.base)]{\protect\node[stlab,rectangle,rounded corners] (X) {steps};}
    and
    \protect\tikz[baseline=(X.base)]{\protect\node[fill=green!25,rectangle,rounded corners] (X) {artefacts};}
    of the proposed method (future work indicated in dashed lines)}
  \label{fig:procedure}
\end{figure}

\Cref{fig:procedure} shows the steps and artefacts of the
proposed method detailed below and
illustrated with a running example.

\subsection{\mgitem[1] Modelling the Manufacturing Process}
\label{sec:activities}

\begin{figure}
\includegraphics[]{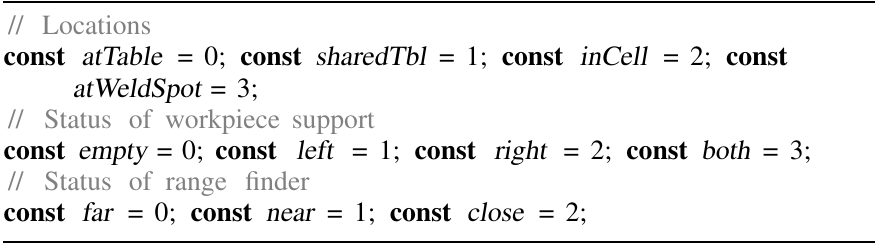}
\caption{Fragment of the data type definition in \textsc{PRISM}\xspace}
\label{lst:statespace}
\vspace{-1.5em}
\end{figure}

Activities in $\mathcal{P}$~(\Cref{fig:setting-activities}) are
structured by sets of guarded commands.
We distinguish \emph{actions} of controllable actors~(e.g.\xspace robot arm,
welder, operator) and the \ac{ASC}, and \emph{events} of a
sensor module and shared ``manipulables''~(e.g.\xspace workpiece
support).
$S$ is built from discrete variables~(cf.\xspace\Cref{lst:statespace})
capturing the world state~(e.g.\xspace robot location; workbench status),
sensory inputs~(e.g.\xspace range finder), control outputs~(e.g.\xspace robot
behaviour, notifications), user inputs~(e.g.\xspace start button), and
modes~(e.g.\xspace current activity, safety mode).

Mode variables~(e.g.\xspace \texttt{ract}, \texttt{safmod}) are used to
specify a \emph{filter} $\phi_a$ for enabling actions that form an
activity~(e.g.\xspace grab workpiece, move arm to welder), or a filter
$\phi_{\mathit{sm}}$ for enabling actions in a particular safety mode.
Thus, the structure of guarded commands for $\mathcal{P}$ follows the
pattern
\[
  [\alpha]\;\neg\omega \land \phi_{\mathit{sm}} \land \phi_a \land
  \gamma \longrightarrow \upsilon
\]
with an action label $\alpha$, a guard $\omega$ to prevent from
leaving the $\mathit{final}$ state, a check $\gamma$ of individual
conditions, and an update expression
$\upsilon$~(cf.\xspace\Cref{sec:prelim}).
Given a set $S_{\mathit{sm}}$ of \emph{safety modes}, modelling
involves the restriction of guarded commands of all actors in
$\mathcal{P}$, by adding $\phi_{\mathit{sm}}$ and $\phi_a$ to their
guards, to obtain mode- and activity-aware guarded
commands. %

\begin{figure}
\includegraphics[]{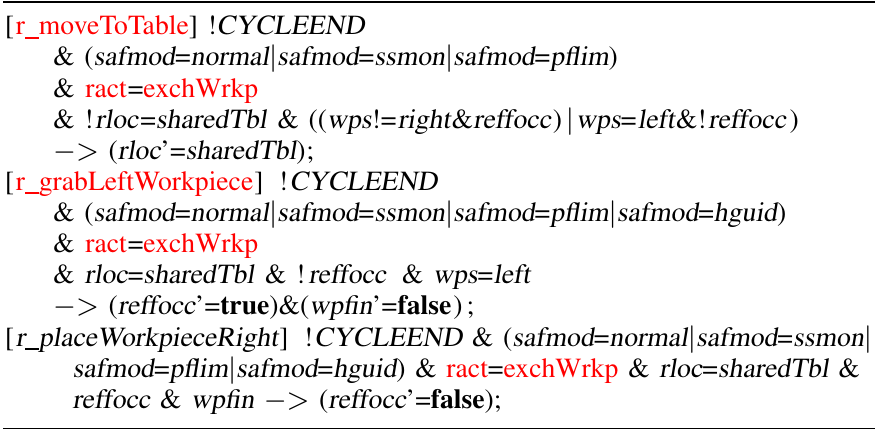}
\caption{\textsc{PRISM}\xspace model fragment of the module \texttt{robotArm}}
\label{lst:robotarm}
\end{figure}
\begin{example}
  \label{exa:robotactions}
  \Cref{lst:robotarm} specifies the two robot actions
  \textcolor{red}{r\_moveToTable} and
  \textcolor{red}{r\_grabLeftWorkpiece} of the activity
  \textcolor{red}{\texttt{exchWrkp}}.
\end{example}

\subsection{\mgitem[2] Safety Analysis and Risk Modelling}
\label{sec:appl-cobots-proj}

\begin{figure}
  \centering
  \footnotesize
\includegraphics[]{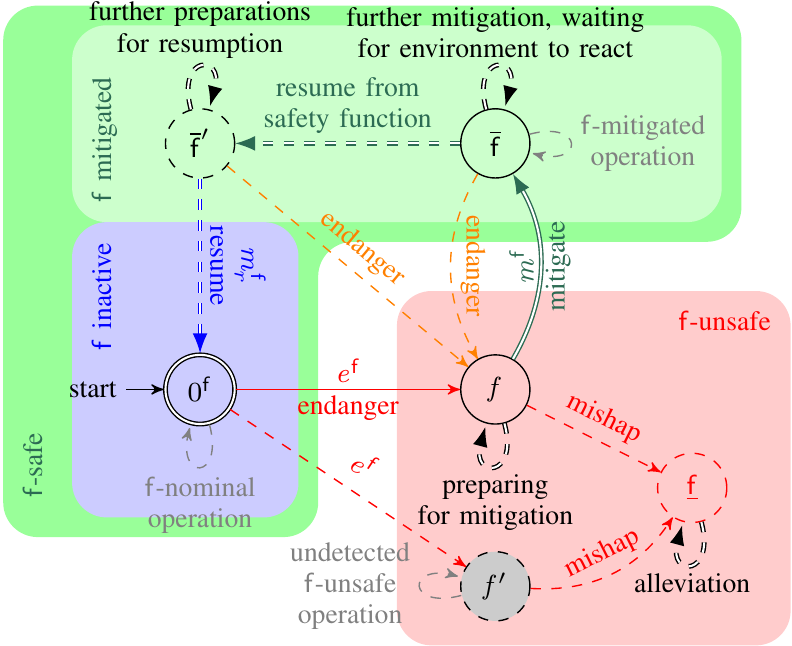}

   \caption{Phases and actions of a risk factor $\rfct$.
    $\Rightarrow$\dots multiple optional actions considered, ---\dots
    minimum amount of information to be provided for a risk factor, -
    -\dots optional modelling aspects.}
  \label{fig:riskfactor}
  \vspace{-1.5em}
\end{figure}

\Cref{fig:riskfactor} further develops the
notion~\cite{Gleirscher2017-FVAV} of a risk factor $\rfct$ towards
guidance in the formalisation of hazards, causes, and mishaps and the
\emph{events} forming a causal chain~(e.g.\xspace a mishap event leads to a
mishap state).  Based on that, $\rfct$ supports the design of hazard
mitigations to reduce accidents, and accident alleviations to reduce
consequences.  Hence, each critical event needs to be translated into
a risk factor.  \Cref{exa:riskfactor} instantiates $\rfct$ with the
hazard~$\rfct[HC]$ from \Cref{tab:riskana}.

\begin{example}
  \label{exa:riskfactor}
  For the hazard~$\rfct[HC]$, \Cref{fig:riskfactor} describes
  \begin{enumerate}[(a)]
  \item how an \emph{endangerment} $\acend{\rfct}$ activates
    $\rfct[HC]$~(i.e.,\xspace leads to a risk state
    $\rho_{\activ[HC]}\in R(F)$ where the predicate $\activ[HC]$ holds
    true),
  \item how \emph{mitigations}~(e.g.\xspace issuing an operator
    notification) update $\mathcal{P}$ to enter the phase $\mitig[HC]$~(i.e.,\xspace
    $\activ[HC]$ is false),
  \item further \emph{mitigations}~(e.g.\xspace waiting for operator
    response),
  \item \emph{resumptions}~(e.g.\xspace switching from speed \& separation
    monitoring to normal) that update $\mathcal{P}$ to return to phase
    $\inact[HC]$ where both $\activ[HC]$ and $\mitig[HC]$ are
    false,
  \item further \emph{endangerments}~(e.g.\xspace erroneous robot movement)
    re-activating $\rfct[HC]$ from state $\mitig[HC]$,
  \item a \emph{mishap} event moving $\mathcal{P}$ into a state with
    $\mishp[HC]$ true~(i.e.,\xspace an $\rfct$-accident occurs),
  \item \emph{alleviations} to handle consequences
    of $\rfct[HC]$ in phase $\mishp[HC]$.
  \end{enumerate}
\end{example}

Phase $\activ'$, reachable by non-deterministic or probabilistic
choice, models an \emph{undetected endangerment}~(e.g.\xspace because of a
faulty range finder for $\rfct[HC]$) that can lead to $\mishp$.  For
sake of simplicity, the $\mathit{endanger}$ choices in $\mitig$ and
$\mitig'$ are not shown.  $\mitig$, $\mitig'$, and $\inact$ form the
\emph{$\rfct$-safe region} of $\mathcal{P}$.  
\Cref{exa:riskfactor:yap} explains how one models risk for the
\texttt{welding} activity in \textsc{Yap}\xspace script.

\begin{figure}
\includegraphics[]{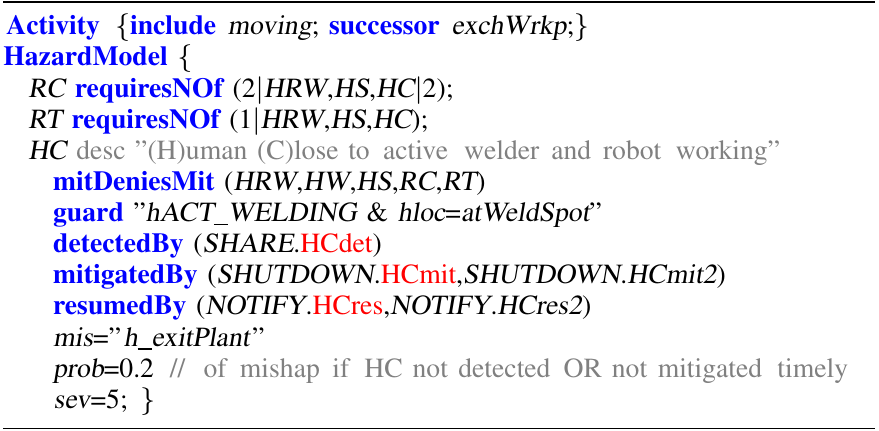}
\caption{\textsc{Yap}\xspace risk model for the \texttt{welding} activity from \Cref{fig:setting-activities}}
\label{lst:moving}
\vspace{-1em}
\end{figure}
\begin{example}
  \label{exa:riskfactor:yap}
  First, the \yapkeyw{Activity} section of \Cref{lst:moving} specifies
  that \texttt{welding} \yapkeyw{includes} the specification of the
  activity \texttt{moving} and that the activity \texttt{exchWrkp} is
  a \yapkeyw{successor} of \texttt{welding}.  This way, one specifies
  an activity automaton for $\mathcal{P}$ as shown in
  \Cref{fig:setting-activities}.

  Next, the \yapkeyw{HazardModel} section lists critical events
  relevant to \texttt{welding}, the two mishaps $\rfct[RC]$ and
  $\rfct[RT]$ and the latent cause $\rfct[HC]$~(cf.\xspace
  \Cref{tab:riskana}).
  One can hypothesise high-level relationships between critical events
  using constraints.  E.g.\xspace $\rfct[RC]$ \yapkeyw{requiresNOf}
  $(2|\rfct[HRW],\rfct[HS],\rfct[HC]|2)$ expresses the assumption that
  exactly two of the listed events have to have occurred before
  $\rfct[RC]$ can occur.  Such relationships are typically identified
  during preliminary \ac{HazOp}, system \acs{FMEA}, or system \acs{FTA}.

  Furthermore, $\rfct[HC]$ is specified by
  \begin{inparaenum}[(a)]
  \item an informal \yapkeyw{desc}ription,
  \item a \yapkeyw{guard} describing its activation $\activ[HC]$,
  \item $mis$~(i.e.,\xspace an action, e.g.\xspace of the operator, with the mishap
    $\mishp[HC]$ as a bad outcome if $\activ[HC]$ is undetected or not
    mitigated timely),
  \item $prob$~(i.e.,\xspace the probability of $\mishp[HC]$ under these
    conditions), and
  \item $sev$, quantifying the severity of the best, average, or worst
    expected consequences from $\mishp[HC]$.
  \end{inparaenum}
\end{example}

\label{sec:uncertainties}
Probabilistic choice in $\mathcal{M}$ can be used to model several
\emph{uncertainties}.
Informed by \ac{FTA} and \ac{FMEA}, one can consider \emph{sensor and
actuator faults}.  In our example, the range finder as the detector of 
$\acend{HC}$ fails by 5\% when the operator enters the cell.
Informed by \ac{HazOp}, \emph{human errors} can be modelled similarly.  
In our example, with a 10\% chance, the operator enters the cell, knowing
that \texttt{robotArm} and \texttt{welder} are active.
Moreover, one can model the probability of occurrence of a
\emph{mishap} under the condition of an active hazard.  In our
example, with a 20\% chance, $\mishp[HC]$ may follow
$\activ[HC']$~(i.e.,\xspace $\rfct[HC]$ remains undetected because of the
aforementioned sensor fault) or $\activ[HC]$~(i.e.,\xspace the \ac{ASC} is not
reacting timely).

\subsection{\mgitem[3] Designing Mitigation and Resumption Options}

\begin{figure}
\includegraphics[]{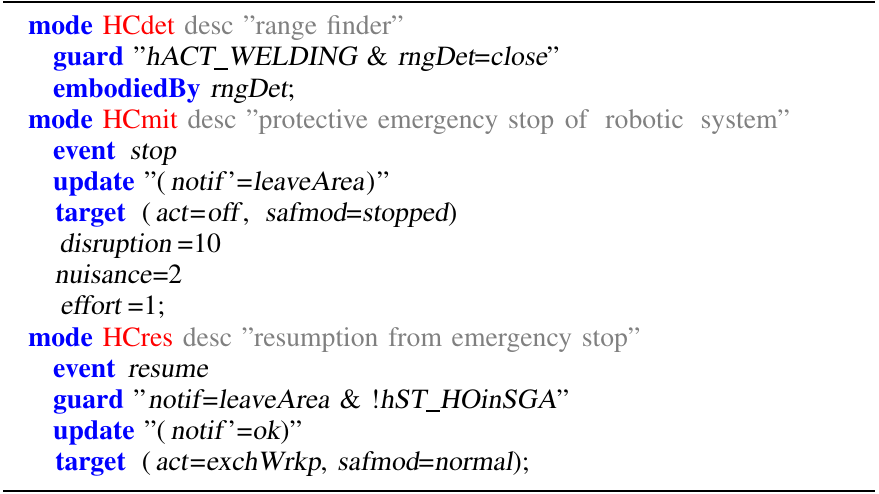}
\caption{\textsc{Yap}\xspace action specifications for the risk factor {$\rfct[HC]$}}
\label{lst:modespec}
\end{figure}

The capabilities of actors in $\mathcal{P}$ determine the
controllability of critical events.  We found three techniques useful
in designing mitigations and resumptions: \emph{action filters}~(i.e.,\xspace
safety modes, cf.\xspace\Cref{sec:intro}),
\emph{activity changes}~(e.g.\xspace change from \texttt{welding} to
\texttt{off}), and \emph{safety functions}~(e.g.\xspace notification).
Recall that mitigations and resumptions are actions~(i.e.,\xspace transition
labels in a risk factor \ac{LTS}).  Accordingly, the example in
\Cref{lst:modespec} specifies details about the actions referred to in
\Cref{lst:moving}.
Here, the following parameters drive the design
space of an \ac{ASC}:
\begin{inparaenum}[(a)]
\item a \yapkeyw{detectedBy} reference~(i.e.,\xspace associating the
  \yapkeyw{guard} with a sensor predicate),
\item a \yapkeyw{mitigatedBy} reference to one or more mitigation
  options, and
\item a \yapkeyw{resumedBy} reference to one or more resumption
  options.
\end{inparaenum}
For this approach, we extended \textsc{Yap}\xspace's input
language to develop these actions into guarded commands.

\begin{example}
  As an example for (b), in \Cref{lst:modespec}, the action
  \textcolor{red}{\textrm{HCmit}} of type \texttt{SHUTDOWN}
  \begin{inparaenum}[(i)]
  \item synchronises with the \texttt{robotArm} and \texttt{welder} on
    the \yapkeyw{event} \texttt{stop},
  \item \yapkeyw{update} models a safety function, issuing a
    notification to the operator to leave the safeguarded area, and
  \item \yapkeyw{target} switches the manufacturing cell to the
    activity \texttt{off} and to the safety mode \texttt{stopped}, all
    triggered by the range finder.
  \end{inparaenum}
\end{example}

\label{sec:optcrit-exa}
Indicated in \Cref{fig:riskfactor}, \textcolor{red}{\textrm{HCmit}}
models one option for $\acmit{\rfct[HC]}$.  One can distinguish
several such options by quantities such as \emph{disruption} of the
manufacturing process, \emph{nuisance} of the operator, and
\emph{effort} %
to be spent by the machines.  In combination with processing
\emph{time} and \emph{value} %
for each nominal action of $\mathcal{P}$, these quantities enable the
evaluation and selection of optimal policies as we shall see below.

This part of the \textsc{Yap}\xspace model can be translated into \ac{pGCL}.
\emph{Endangerments} are translated into commands of the form
\[
  [\acend{\rfct'}]\;\phi_a \land \chi \longrightarrow
  \activ'
  \quad\text{and}\quad
  [\acend{\rfct}]\;\phi_a \land \zeta \longrightarrow
  (1-p)\colon\activ + p\colon\activ'
\]
with guards including a hazard condition $\chi$ and a corresponding
monitoring~(or sensor) predicate $\zeta$.  Constraints, such as
\yapkeyw{requiresNOf} in \Cref{exa:riskfactor:yap}, are then used to
derive part of $\zeta$.

\begin{figure}
\includegraphics[]{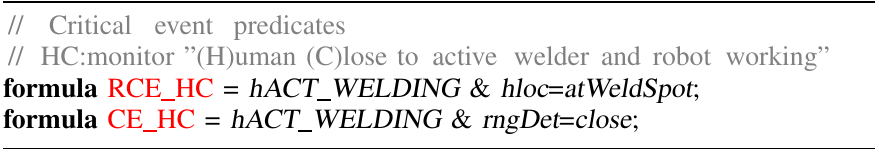}
\caption{Monitor predicates for {$\rfct[HC]$} generated for the \textsc{PRISM}\xspace model}
\label{lst:monitors}
\vspace{-1.5em}
\end{figure}

\begin{example}
  \Cref{lst:monitors} indicates the transcription of \yapkeyw{guard}
  and \yapkeyw{detectedBy} into a pair of predicates,
  $\mathit{RCE\_HC}$ describing actual states, and $\mathit{CE\_HC}$
  signifying states monitored by the range finder, where $p$ can denote
  the sensor fault probability.
\end{example}

\emph{Mitigations} are translated into commands of the form
\[
  [\acmit[t]{\rfct}]\;\phi_t \land \activ \longrightarrow \upsilon_{t'}
  \quad\text{and}\quad
  [\acmit{\rfct}]\;\phi_{\mathit{sm}',a',\mathit{sf}'} \land \activ
  \longrightarrow \mitig 
\]
with $t \in\{\mathit{sm},a,\mathit{sf}\}$ in $\phi_t$ for checking
permission in the current safety mode, activity, and state of safety
functions, and in $\upsilon_{t'}$ for hazard removal by
switching into a safer activity $a'$, a safer mode $\mathit{sm}'$, and
by applying the safety function $\mathit{sf}$.  These updates are
checked by $\phi_{\mathit{sm}',a',\mathit{sf}'}$ to be able to proceed
to $\mitig$.
\emph{Resumptions} are translated into commands of the form
\[
  [\acmit[r,t]{\rfct}]\;\phi_t \land \rho_{\mitig} \longrightarrow
  \upsilon_{t'}
  \quad\text{and}\quad
  [\acmit[r]{\rfct}]\;\phi_{\mathit{sm}',a',\mathit{sf}'} \land
  \rho_{\mitig} \longrightarrow \inact
\]
where $\phi_t$ guards the resumption based on the safety mode and
function in place, $\rho_{\mitig}\subseteq R(F)$ restricts permission
to risk states~(\Cref{sec:yap} and \Cref{fig:riskfactor}) with $\rfct$
mitigated; and $\upsilon_{t'}$ inverts the safety
function~($\mathit{sf}^{-1}$), relaxes to the safety mode
$\mathit{sm}'$, and returns to an, ideally more productive, activity
$a'$ of $\mathcal{P}$.

\subsection{\mgitem[4] Verified Controller Synthesis}
\label{sec:verif}

The present approach follows a two-staged search through the
\ac{ASC} design space: The first stage is carried through by \textsc{Yap}\xspace when
generating the guarded commands.  The second stage is performed by
\textsc{PRISM}\xspace when synthesising \ac{MDP} policies.  For search space
reduction, \textsc{Yap}\xspace employs \emph{risk gradients} between safety modes and
activities in the first stage.  For the second stage, \textsc{Yap}\xspace generates
reward structures for some of the quantities introduced in
\Cref{sec:optcrit-exa}.

\subsubsection{Guarded Command Generation}

The generation of $\upsilon_{t'}$ for mitigations and resumptions
requires the choice of a safety mode and activity to switch
to, depending on the current mode and activity.  Given
activities $S_a$ and modes $S_{\mathit{sm}}$, two skew-diagonal
\emph{risk gradient matrices}
$\mathfrak{S}^a\in\mathbb{R}^{|S_a|\times|S_a|}$ and
$\mathfrak{S}^{\mathit{sm}}\in\mathbb{R}^{|S_{\mathit{sm}}|\times|S_{\mathit{sm}}|}$,
e.g.\xspace manually crafted from safety analysis,
can resolve this choice based on the following justification.

Assume $a_1, a_2\in S_a$ vary in physical movement, force, and speed.
If $a_1$ means more or wider movement, higher force application, or
higher speed than $a_2$, then a change from $a_1$ to $a_2$ will
likely reduce risk.  Hence, a positive gradient is assigned to
$\mathfrak{S}^{a}_{a_1 a_2}$.
Similarly, assume $m_1, m_2 \in S_{\mathit{sm}}$ vary $\mathcal{P}$'s
capabilities by relaxing or restricting the range and shape of
permitted actions.  If $m_1$ permits stronger capabilities than $m_2$,
then a change from $m_1$ to $m_2$ will likely reduce risk.  Again, we
assign a positive gradient to $\mathfrak{S}^{\mathit{sm}}_{m_1 m_2}$.
The diagonality of $\mathfrak{S}$ provides the dual for resumptions
where a negative gradient of the same amount from $m_2$ to $m_1$ is
assigned to $\mathfrak{S}^{\mathit{sm}}_{m_2 m_1}$.

\begin{figure}
\includegraphics[]{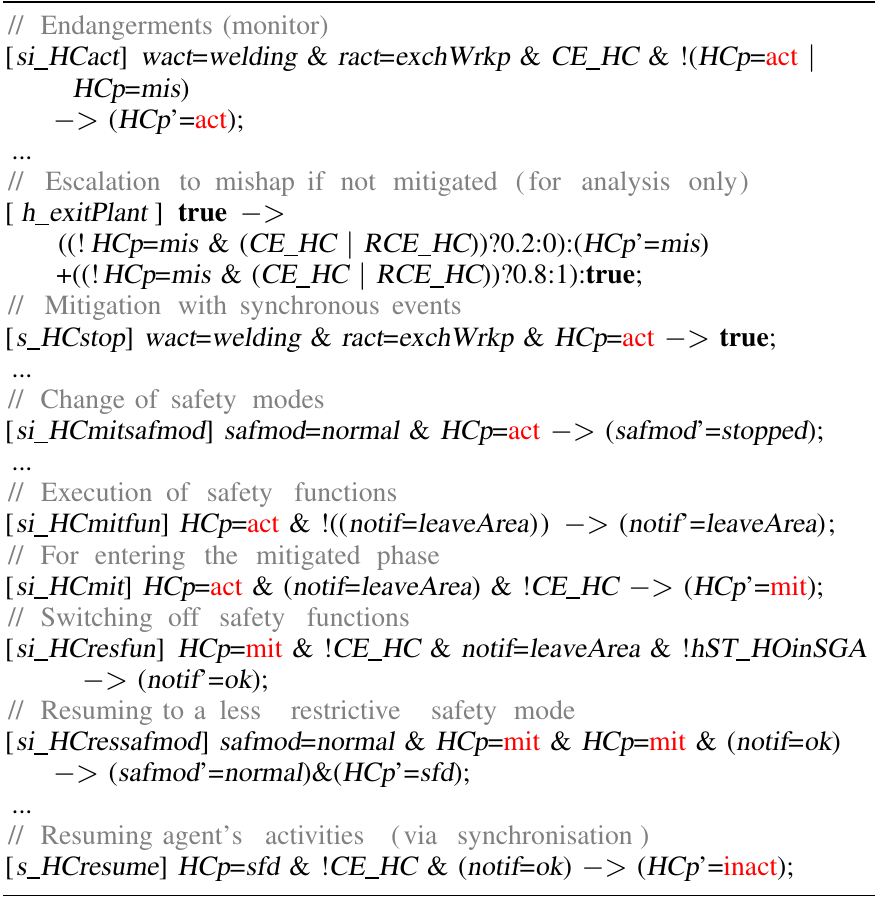}
\caption{\textsc{PRISM}\xspace model fragment generated for the risk factor {$\rfct[HC]$}}
\label{lst:mitres}
\end{figure}

Let current safety mode $c$ and mitigation $\acmit{\rfct}$ with
\yapkeyw{target} mode $t$.
$\acmit{\rfct}$ ($\acmit[r]{\rfct}$) changes to $t$ only if the gradient
from $c$ to $t$ is $\geq 0$ ($\leq 0$).  If
$\mathfrak{S}^{\mathit{sm}}_{c t} \geq 0$,
then a switch to $t$ is included in $\upsilon_{\mathit{sm}'}$, otherwise
$\upsilon_{\mathit{sm}'}$ leaves $\mathcal{M}$ in $c$.  We
implemented this scheme for activities analogously.
\begin{example}
  \Cref{lst:mitres} shows the result of applying this scheme in the
  generation of an \ac{ASC} for the risk factor $\rfct[HC]$ based on
  \Cref{lst:modespec}.
\end{example}
Essentially, $\mathfrak{S}$ approximates the change of risk in case of
a change from one activity or safety mode to another.  Using
$\mathfrak{S}$, the majority of an \ac{ASC} can be described in \textsc{Yap}\xspace
script.

\subsubsection{\ac{MDP} Verification}
\label{sec:verif-mdp}

This step requires establishing
$\mathcal{M} \models \phi_{\mathit{wf}} \land \phi_c$ with properties
expressed in \ac{PCTL}~(\Cref{sec:prelim}).  $\phi_{\mathit{wf}}$ is a
\emph{well-formedness} property including the verification of, e.g.\xspace
hazard occurrence and freedom of pre-$\mathit{final}$ deadlocks, and
the falsification, e.g.\xspace that final states must not be initial states.
$\phi_{\mathit{wf}}$ helps to simplify model debugging, decrease model
size, guarantee progress, and reduce vacuity.  $\phi_c$ specifies
\emph{safety-carrying correctness} including, e.g.\xspace \ac{ASC}
progress~(and across cycles, liveness), particularly, that
$\Pi_{\mathcal{M}}$~(i.e.,\xspace the \ac{ASC} design space) contains complete
mitigation paths from critical events.  \Cref{tab:properties} lists
examples of $\phi_{\mathit{wf}}$ and $\phi_c$ to be verified of
$\mathcal{M}$.

\subsubsection{Policy Synthesis}

The \ac{ASC} design space ($\Pi_{\mathcal{M}}$) is created by 
commands~(e.g.\xspace mitigations, resumptions) simultaneously enabled in 
$s\in S$, yielding multiple policies for $s$ and some 
commands enabled in multiple states, giving rise to a policy for
each ordering in which these commands can be chosen.

An \emph{optimal policy} $\pi^{\star}$, including the \ac{ASC}
decisions, can be selected from $\Pi_{\mathcal{M}}$ based on multiple
criteria~(e.g.\xspace minimum risk and nuisance, maximum productivity).  For
that, $\mathcal{M}$ uses action rewards to quantify
\begin{inparaenum}[(i)]
\item \textit{prod}uctivity, 
up- and %
down-time of $\mathcal{P}$; %
\item factor-, %
mode-, and activity-based \textit{risk}; %
risk reduction \textit{pot}ential; %
disruptiveness and \textit{nuis}ance; %
resource consumption; and %
\textit{eff}ective time of the \ac{ASC}. %
\end{inparaenum}

\begin{figure}
\includegraphics[]{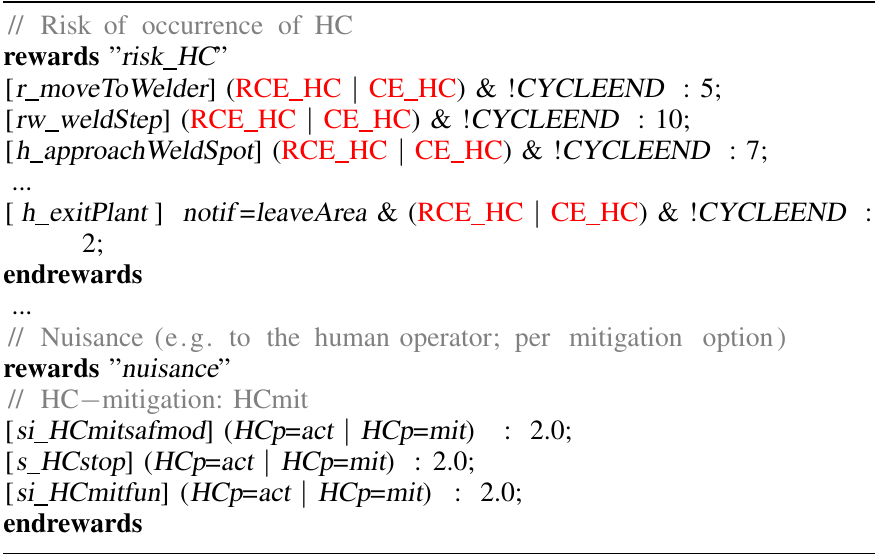}
\caption{\textsc{PRISM}\xspace rewards for risk from {$\rfct[HC]$} and nuisance of \texttt{HCmit}}
\label{lst:rewards}
\end{figure}

\begin{example}
  \Cref{lst:rewards} shows a \ac{pGCL} fragment generated by \textsc{Yap}\xspace.
\end{example}

\subsubsection{\Ac{DTMC} Verification}
\label{sec:verif-dtmc}

Due to known restrictions in combining multi-objective queries and
constraints in \textsc{PRISM}\xspace, part of the verification applies to the policy
as a \ac{DTMC}.  This step requires establishing
$\pi^{\star}\models\phi_s$ where $\phi_s$ can include liveness,
safety, and reliability properties~(e.g.\xspace ``reach-avoid'' of type
$\mathop{\textbf{A}}\mathop{\textbf{G}}\mathop{\textbf{F}}\psi \land
\mathop{\textbf{A}}\mathop{\textbf{G}}\neg\phi$; the probability of failure on
demand of the \ac{ASC}; the probability of a mishap from any hazard is
below a threshold).  \Cref{tab:properties} lists examples of
properties to be verified of $\pi^{\star}$.

\begin{table}[t]
  \centering
  \caption{Examples of checked properties and queried objectives}
  \label{tab:properties}
  \footnotesize
  \begin{tabularx}{\columnwidth}{%
    >{\hsize=.5\hsize}X
    >{\hsize=.5\hsize}X}
    \toprule
    \textbf{Property}$^\dagger$
    & \textbf{Description}
    \\\specialrule{\lightrulewidth}{\aboverulesep}{0em}
    \multicolumn{2}{l}{\cellcolor{lightgray}{\large$\phantom{\mid}$\hspace{-.25em}}\emph{Well-formedness $\phi_{\mathit{wf}}$ of $\mathcal{M}$}}
    \\\specialrule{\lightrulewidth}{0em}{\belowrulesep}
    v: $\mathop{\textbf{E}}\mathop{\textbf{F}}(\activ\land\neg \mathit{final})$ %
    & Can the hazard $\rfct$ occur during a cycle of $\mathcal{P}$?
    \\
    f: $\mathop{\textbf{E}}\mathop{\textbf{F}}(\mathit{deadlock} \land\neg \mathit{final})$
    & Are all deadlock states final? Does $\mathcal{P}$ deadlock
    early?
    \\
    f: $\mathop{\textbf{A}}\mathop{\textbf{F}}\activ$ %
    & Is $\rfct$ inevitable?
    \\
    f: $\neg\exists s\in S\colon \mathit{final} \land \mathit{init}$
    & Are there initial states that are also final states?
    \\
    v: $\mathop{\textbf{E}}\mathop{\textbf{F}} \mathit{final}$ 
    & Can $\mathcal{P}$ finish a production cycle?
    \\\specialrule{\lightrulewidth}{\aboverulesep}{0em}
    \multicolumn{2}{l}{\cellcolor{lightgray}{\large$\phantom{\mid}$\hspace{-.25em}}\emph{Querying for a (Pareto-)optimal \ac{ASC}
        $\pi^{\star}$}}
    \\\specialrule{\lightrulewidth}{0em}{\belowrulesep}
    $\mathop{\textbf{R}}^{\mathit{pot}}_{\max=?} [ \mathop{\textbf{C}} ] \land
    \mathop{\textbf{R}}^{\mathit{eff}}_{\max=?} [ \mathop{\textbf{C}} ]$ 
    & Assuming an adversarial environment, select
    $\pi$ that maximally utilises the \ac{ASC}. 
    \\
    $\mathop{\textbf{R}}^{\mathit{prod}}_{\max=?} [ \mathop{\textbf{C}} ] \land \mathop{\textbf{R}}^{\mathit{sev}}_{\leq s}
    [ \mathop{\textbf{C}} ] \land \mathop{\textbf{R}}^{\mathit{risk}}_{\leq r} [ \mathop{\textbf{C}} ]$
    & Select \ac{ASC} that maximises productivity 
    constrained by risk level $r$ and expected severity $s$. 
    \\
    $\mathop{\textbf{R}}^{\mathit{prod}}_{\max=?} [ \mathop{\textbf{C}} ] \land \mathop{\textbf{R}}^{\mathit{sev}}_{\leq
      s} [ \mathop{\textbf{C}} ]$ 
    & Select \ac{ASC} that maximises productivity constrained by
    exposure $p$ to severe injuries. 
    \\\specialrule{\lightrulewidth}{\aboverulesep}{0em}
    \multicolumn{2}{l}{\cellcolor{lightgray}{\large$\phantom{\mid}$\hspace{-.25em}}\emph{Cycle-bounded correctness
        $\phi_c$ of a policy $\pi$ (or the policy space $\Pi_{\mathcal{M}}$)}}
    \\\specialrule{\lightrulewidth}{0em}{\belowrulesep}
    v: $\mathop{\textbf{A}}\mathop{\textbf{F}} (\zeta \rightarrow \mathop{\textbf{A}}\mathop{\textbf{X}} \activ)$
    & Does the \ac{ASC} on all paths immediately detect the hazard $\chi$?
    \\
    v: $\mathop{\textbf{A}}\mathop{\textbf{F}} (\activ \rightarrow (\mathop{\textbf{A}}\mathop{\textbf{F}}
    \mitig \rightarrow (\mathop{\textbf{A}}\mathop{\textbf{F}}\inact )))$
    & Does the \ac{ASC} lively handle hazard $\rfct$ in all
    situations?
    \\
    v: $\mathop{\textbf{E}}\mathop{\textbf{F}}(\activ \land \mathop{\textbf{F}} \mathit{final})$
    & Does the \ac{ASC} resume $\mathcal{P}$ so it can finish its
    cycle after $\rfct$ has occurred?
    \\
    v: $\mathop{\textbf{P}}_{>p}[\mathop{\textbf{G}} \neg \mathit{mishap}]$
    & Is the probability of mishap freedom greater than $p$?
    \\\specialrule{\lightrulewidth}{\aboverulesep}{0em}
    \multicolumn{2}{l}{\cellcolor{lightgray}{\large$\phantom{\mid}$\hspace{-.25em}}\emph{Reliability $\phi_r$ of a selected
        \ac{ASC} $\pi^{\star}$}} 
    \\\specialrule{\lightrulewidth}{0em}{\belowrulesep}
    v: $\mathop{\textbf{S}}_{< p} \mathit{mishap}$
    & Is the steady-state (long-run) probability of any mishap
    $\mishp$ below $p$?
    \\\bottomrule
    \multicolumn{2}{p{.95\columnwidth}}{
      $^{\dagger}$
      $\mathit{deadlock}$ \dots state with no commands enabled,
      $\mathit{final}$ \dots end of manufacturing cycle,
      $\mathit{init}$ \dots initial state of a manufacturing cycle,
      $\mathit{mishap}$ \dots mishap state,
      $p$ \dots probability bound,
      v \dots to be verified,
      f \dots to be falsified,
      $\mathit{prod}$ \dots productivity,
      $\mathit{sev}$ \dots severity,
      $\mathit{eff}$ \dots \ac{ASC} effectiveness,
      $\mathit{risk}$ \dots risk level,
      $\mathit{pot}$ \dots risk reduction potential
    }
  \end{tabularx}
\end{table}

\begin{figure}
\includegraphics[]{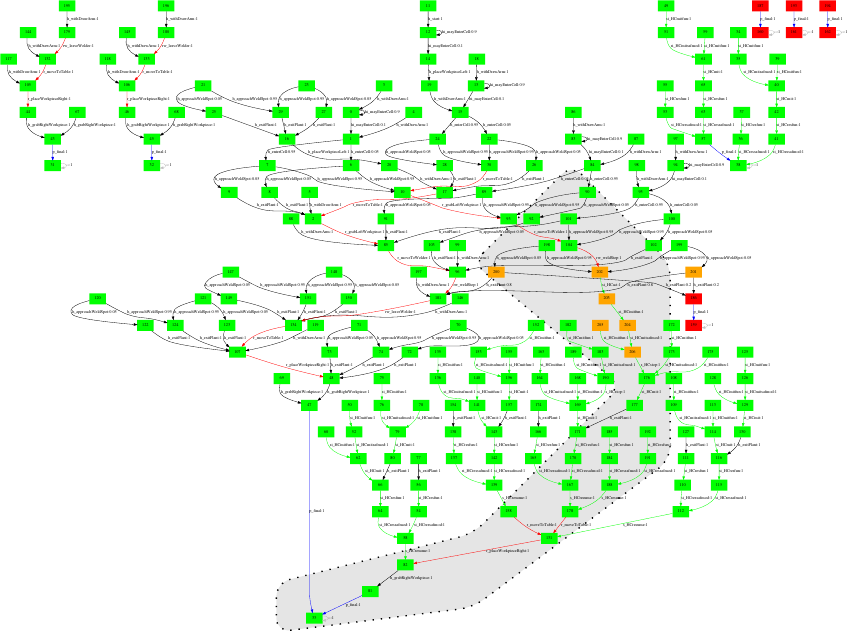}
  \caption{Bird's-eye view of the policy synthesised for the query
    $\mathop{\textbf{R}}^{\mathit{pot}}_{\max=?} [ \mathop{\textbf{C}} ]
    \land \mathop{\textbf{R}}^{\mathit{eff}}_{\max=?}
    [ \mathop{\textbf{C}} ]$.  Nodes are the states
    reachable in $\mathcal{M}$ from $s_0$, including $\rfct[HC]$-safe states
    (green), $\rfct[HC]$-unsafe states (orange), and mishap states (red).  Edges indicate
    \texttt{robotArm} and \texttt{welder} actions (red), actions of
    the operator (black), the \ac{ASC} (green), and cycle
    termination (blue).
    The gray fragment is magnified in \Cref{fig:synctr-cutout}.
    \label{fig:synctr-all}}
\end{figure}

\Cref{fig:synctr-all} visualises $\pi^{\star}$ as a graph with nodes
for all states reachable from $s_0$, %
and edges for all transitions generated by $\delta_{\mathcal{P}}$ as
derived from the guarded commands of $\mathcal{P}$.  The edges form
executions of $\mathcal{M}$ from $s_0$ under $\pi^{\star}$.

\begin{figure}
  \centering
  \includegraphics[width=.6\columnwidth]{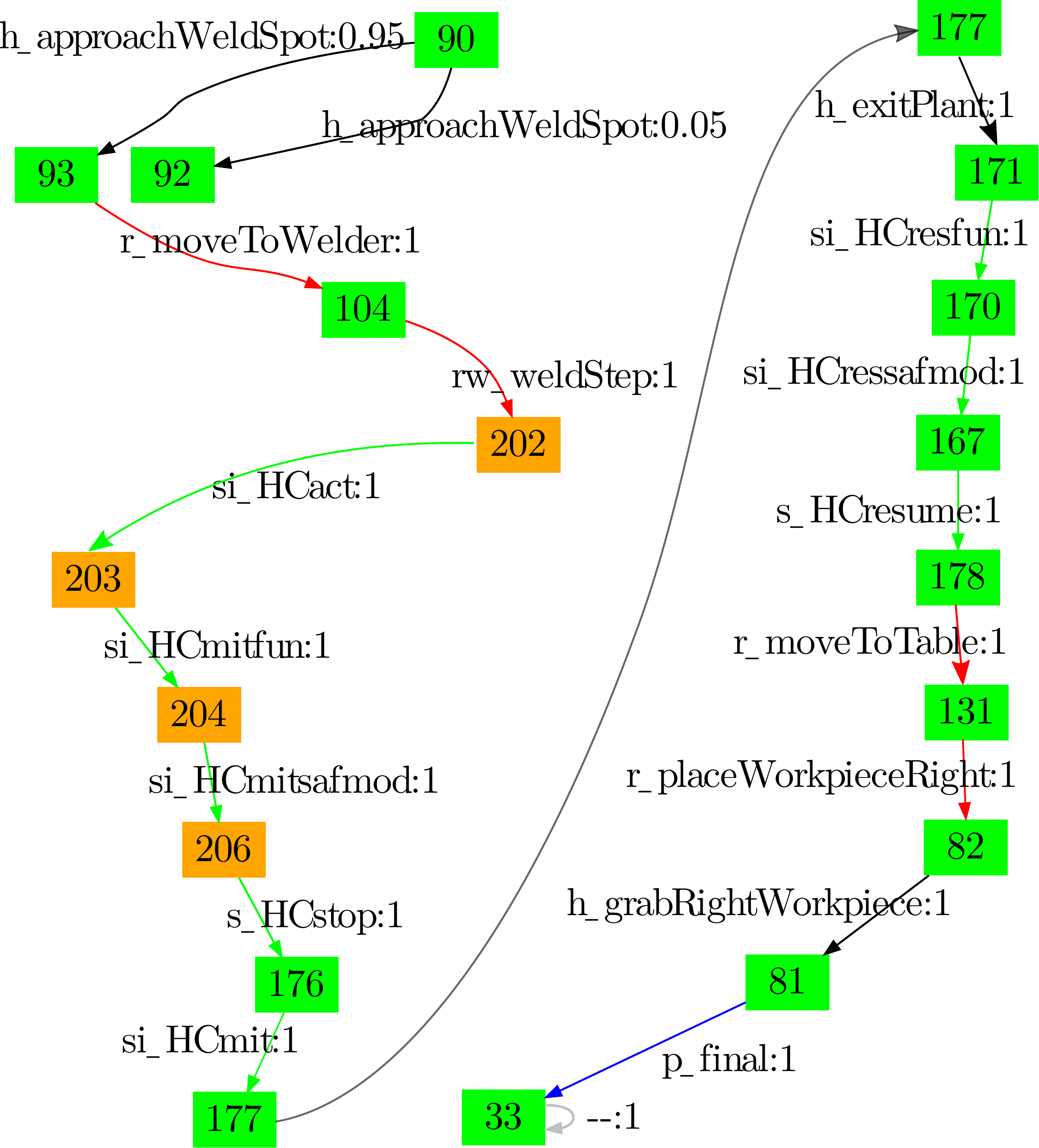}
  \caption{Fragment of the policy shown in
    \Cref{fig:synctr-all}.  Split at state 177 for layout efficiency.
    \label{fig:synctr-cutout}}
\end{figure}

\begin{example}
  \Cref{fig:synctr-all} provides a bird's eye view of a synthesised
  policy.  States coloured in green form the set $\rfct[HC]$-safe,
  i.e.,\xspace states in which $\rfct[HC]$ is inactive or mitigated.

  \Cref{fig:synctr-cutout} shows a fragment of \Cref{fig:synctr-all}.
  Note the 5\% chance of a sensor failure from state 90 leading to
  state 92~(i.e.,\xspace the \texttt{operator} has approached the
  \texttt{robotArm} and \texttt{welder}) where $\rfct[HC]$ will remain
  undetected and not handled.  Otherwise, in state 93, $\rfct[HC]$
  will be mitigated after the next work step of the \texttt{robotArm} and
  \texttt{welder} leading to state 202.  From there, the \ac{ASC}
  mitigates to a protective \texttt{stop}~(state 176), and resumes to
  state 178~(i.e.,\xspace where the \texttt{operator} has left the safeguarded
  area) from where the current manufacturing cycle can be
  finished~(state 33).
\end{example}

\section{Evaluation}
\label{sec:eval}

In this section, we discuss the adequacy and efficacy of the proposed
method from several viewpoints.

\subsection{Research Questions and Evaluation Methodology}
\label{sec:methodology}

Based on the questions raised in \Cref{sec:safety-req}, we
investigate the
\begin{inparaenum}[(i)]
\item scalability and performance of the approach and
\item the effectiveness of the \acp{ASC} synthesised by it,
\end{inparaenum}
asking:
\begin{description}
\item[RQ1] How well can the approach deal with multiple hazards
  and mitigation and resumption options?
  What are the resulting model sizes and analysis times?
\item[RQ2] What is the likelihood of incident/accident-free operation 
  under the control of the synthesised \acp{ASC}?
\item[RQ3] Which process overheads are to be expected of an \ac{ASC}
  implementation?
\end{description}

For \textbf{RQ1}, we consider as inputs and parameters a \textsc{Yap}\xspace risk
model and a \textsc{PRISM}\xspace \ac{MDP} model of the cell (with \textsc{Yap}\xspace template
placeholders), a single initial state of these models where all actors are in
the activity \texttt{off} and no hazard is active.
Accordingly, we prepare and analyse multiple increments of the risk
model, each adding one critical event, mitigation options, and
constraints to the model.

For \textbf{RQ2}, let $\Xi\subset S$ be the set of non-accident
$F$-unsafe states, i.e.,\xspace states labelled with at least one critical
event, describing the abstract state where any critical event has at
least been sensed by the \ac{ASC}~(e.g.\xspace $\mathit{CE\_HC}$ with its
handling not yet started, i.e.,\xspace $\inact[HC]$).  For \acp{MDP}, we
evaluate accident freedom with
\begin{align}
  \label{eq:accfreedom}
  {\mathop{\textbf{P}}}_{\neg A}\equiv
  f_{s\in\Xi}\; {\mathop{\textbf{P}}}^s_{\min=?}
  [\neg \mathit{mishap} \mathop{\textbf{W}} \mathit{safe}]
\end{align}
where $f\in\{\min, \mathrm{mean}, \max\}$.  For $\Xi$,
\Cref{eq:accfreedom} requires the \ac{ASC} to minimise the probability
of mishaps until an $F$-safe state (i.e.,\xspace $S\setminus\Xi$) is reached.
In \Cref{tab:exp:results}, $\lfloor\mu\rceil$ denotes the triple
comprising $\min$, the arithmetic $\mathrm{mean}\;\mu$, and $\max$.
$\mathop{\textbf{P}}_{\neg A}$ aggregates these three probabilities over $\Xi$.

Next, we synthesise policies for each of the \ac{MDP} increments for
the three optimisation queries
\begin{align}
  \label{eq:opt_a}
  {\mathop{\textbf{R}}}^{\mathit{pot}}_{\max=?} [ \mathop{\textbf{C}} ]
  &\land {\mathop{\textbf{P}}}_{\max=?} [ \mathop{\textbf{F}} \mathit{final}_{\mathit{t}}],
    \tag{a}
  \\
  \label{eq:opt_b}
  {\mathop{\textbf{R}}}^{\mathit{prod}}_{\max=?} [ \mathop{\textbf{C}} ]
  &\land {\mathop{\textbf{P}}}_{\max=?} [ \mathop{\textbf{F}}
    \mathit{final}_{\mathit{t}}], \;\text{and}
    \tag{b}
  \\
  \label{eq:opt_c}
  {\mathop{\textbf{R}}}^{\mathit{eff}}_{\max=?} [ \mathop{\textbf{C}} ]
  &\land {\mathop{\textbf{R}}}^{\mathit{nuis}}_{\max=?} [ \mathop{\textbf{C}} ].
    \tag{c}
\end{align}
where $\mathit{final}_t = \{s\in S\mid s\in\mathit{final} \land
\mbox{all tasks finished}\}$. %
In the spirit of negative testing, \Cref{eq:opt_a} aims at maximising
the use of the \ac{ASC}~(i.e.,\xspace approximating worst-case behaviour of
the operator and other actors) while maximising the probability of
finishing two tasks, i.e.,\xspace finishing a workpiece and carrying through
cell maintenance.  This query does not take into account further
opmitisation parameters defined for mitigations and resumptions.
As opposed to that, \Cref{eq:opt_b} fosters the maximisation of
\textit{prod}uctivity, any combination of decisions allowing the
finalisation of tasks is preferred, hence, transitions leading to
accidents or the use of the \ac{ASC} are equally neglected.
While \Cref{eq:opt_c} also forces the environment to trigger the
\ac{ASC}, these policies represent the best \ac{ASC} usage in terms of
\textit{nuis}ance and \textit{eff}ort.  Because of constraints in the
use of $\mathop{\textbf{R}}_{\min}$ for \acp{MDP}, we maximise costs interpreting
positive values as negative~(e.g.\xspace the higher the nuisance the better).

We investigate the Pareto curves of the policies synthesised from the
\Cref{eq:opt_a,eq:opt_b,eq:opt_c}.  For policies with less than 1000
states, we inspect the corresponding policy graphs~(e.g.\xspace whether there
is a path from $\mathit{initial}$ to $\mathit{final}$ or whether paths
from unsafe states reachable from $\mathit{initial}$ avoid deadlocks).
Finally, we evaluate accident freedom according to
\Cref{eq:accfreedom}, except that we use $\mathop{\textbf{P}}_{=?}$ for \acp{DTMC}
instead of $\mathop{\textbf{P}}_{\min=?}$.\footnote{To keep manual workload under
  control, if \textsc{PRISM}\xspace lists several adversaries, we apply the
  experiment procedure only to the first listed.}

\subsection{Results}

\begin{table*}
  \centering
  \caption{Results of the experiment for \textbf{RQ1}~(scalability)
    and \textbf{RQ2}~(accident-free operation)}
  \label{tab:exp:results}
  \footnotesize
  \setlength{\tabcolsep}{5.5pt}
  \begin{tabularx}{\textwidth}{Xcrr|crc|crr|crr|crr}
    \toprule
    \multicolumn{4}{c|}{\textbf{Risk Model}$^\dagger$} 
    & \multicolumn{3}{c|}{\textbf{\ac{MDP}}$^\dagger$}
    & \multicolumn{3}{c}{\textbf{(a) max-ASC}$^\dagger$}
    & \multicolumn{3}{c}{\textbf{(b) max-prod}}
    & \multicolumn{3}{c}{\textbf{(c) opt-ASC}}
    \\\rowcolor{lightgray}{\large$\phantom{\mid}$\hspace{-.25em}}
    $F$ & $mr/c$ & $\vert R(F)\vert$ & $t_Y$
    & $\mathop{\textbf{P}}_{\neg A}$  & $\Xi$ & $sta/tra$
    & $\mathop{\textbf{P}}_{\neg A}$ & $\Xi$ & $t_P$
    & $\mathop{\textbf{P}}_{\neg A}$ & $\Xi$ & $t_P$
    & $\mathop{\textbf{P}}_{\neg A}$ & $\Xi$ & $t_P$
    \\\rowcolor{lightgray}{\large$\phantom{\mid}$\hspace{-.25em}}
    & & & [ms] & $\lfloor\mu\rceil$ & &
    & $\lfloor\mu\rceil$ & & [s]
    & $\lfloor\mu\rceil$ & & [s]
    & $\lfloor\mu\rceil$ & & [s]
    \\\specialrule{\lightrulewidth}{0em}{\belowrulesep} %
    $\rfct[HC]$ & 5/0 & 3 & 40 & [.9,.9,.9] & 14 & 322/1031 %
    & [1,1,1] & 3 & .02
    & [1,1,1] & 1 & .02
    & [1,1,1] & 6 & .15
    \\ %
    $+\rfct[HS]$ & 9/2 & 5 & 52 & [.92,.96,.98] & 256 & 930/3483 %
    & [.07,.66,1] & 11 & .77
    & [0,.88,1] & 8 & .82
    & [.95,.98,1] & 18 & .9 %
    \\ %
    $+\rfct[WS]$ & 11/3 & 8 & 44 & [.93,.97,1] & 288 & 1088/3865 %
    & [0,.29,1] & 17 & 2.1
    & [0,.8,1] & 5 & 2 
    & [1,1,1] & 24 & 1.5
    \\ %
    $+\rfct[HRW]$ & 13/7 & 16 & 65 & [.93,.97,1] & 981 & 7675/33322
    & [1,1,1] & 17 & 9.7
    & [1,1,1] & 11 & 9.4 
    & [1,1,1] & 15 & 13.3 
    \\ %
    $+\rfct[HW]$ & 15/8 & 36 & 76 & [.93,.97,1] & 2296 & 21281/98694
    & [1,1,1] & 15 & 42.9
    & [0,.71,1] & 7 & 41.4 
    & [1,1,1] & 15 & 46.6
    \\ %
    $+\rfct[RT]$ & 15/9 & 50 & 87 & [.93,.97,1] & 2864 & 21965/100133
    & [1,1,1] & 13 & 48.2
    & [1,1,1] & 9 & 46.4
    & [1,1,1] & 15 & 53.8
    \\ %
    $+\rfct[RC]$ & 15/15 & 122 & 162 & [.93,.99,1] & 12079 & 21670/102263
    & [0,.94,1] & 35 & 38
    & [0,.72,1] & 22 & 36.6
    & [1,1,1] & 36 & 51.1
    \\\bottomrule
    \multicolumn{16}{p{.98\textwidth}}{%
      $^{\dagger}$
      $F$\dots critical event set;
      mr/c\dots number of mitigations+resumptions/constraints;
      $\vert R(F)\vert$\dots cardinality of the risk space;
      $t_Y$\dots \textsc{Yap}\xspace's processing time;
      $\mathop{\textbf{P}}_{\neg A}$\dots probability of conditional accident freedom;
      $\Xi$\dots set of $F$-unsafe states;
      $sta/tra$\dots number of states/transitions of the \ac{MDP} ($sta$ equals
      the size of the policies);
      \Cref{eq:opt_a,eq:opt_b,eq:opt_c}\dots optimisation queries;
      $t_P$\dots \textsc{PRISM}\xspace's processing time
    }
\end{tabularx}
\end{table*}

For the experiment, we used \textsc{Yap}\xspace~0.5.1 and \textsc{PRISM}\xspace~4.5, on
GNU/Linux~5.4.19 (x86, 64bit), and an Intel\textregistered~Core
i7-8665U with up to 8~CPUs of up to 4.8~MHz, and 16~GiB RAM.

\Cref{tab:exp:results} shows the data collected from seven models
created for \textbf{RQ1} and \textbf{RQ2}.
The result $\lfloor\mu\rceil = [1,1,1]$ for a policy denotes 100\%
conditional accident freedom.  This desirable result is most often
achieved with \Cref{eq:opt_c} due to the fact that simultaneity of
decisions of the environment and the \ac{ASC} in the same state is
avoided by focusing on rewards only specified for \ac{ASC} actions.
Such rewards model the fact that an \ac{ASC} is usually much faster
than an operator.
\Cref{eq:opt_a,eq:opt_b} show poorer accident freedom because
\textit{prod}uctivity rewards given to the environment compete with
rewards given to the \ac{ASC} to exploit its risk reduction
\textit{pot}ential.

\begin{figure}
  \centering
  \includegraphics[width=.7\columnwidth]{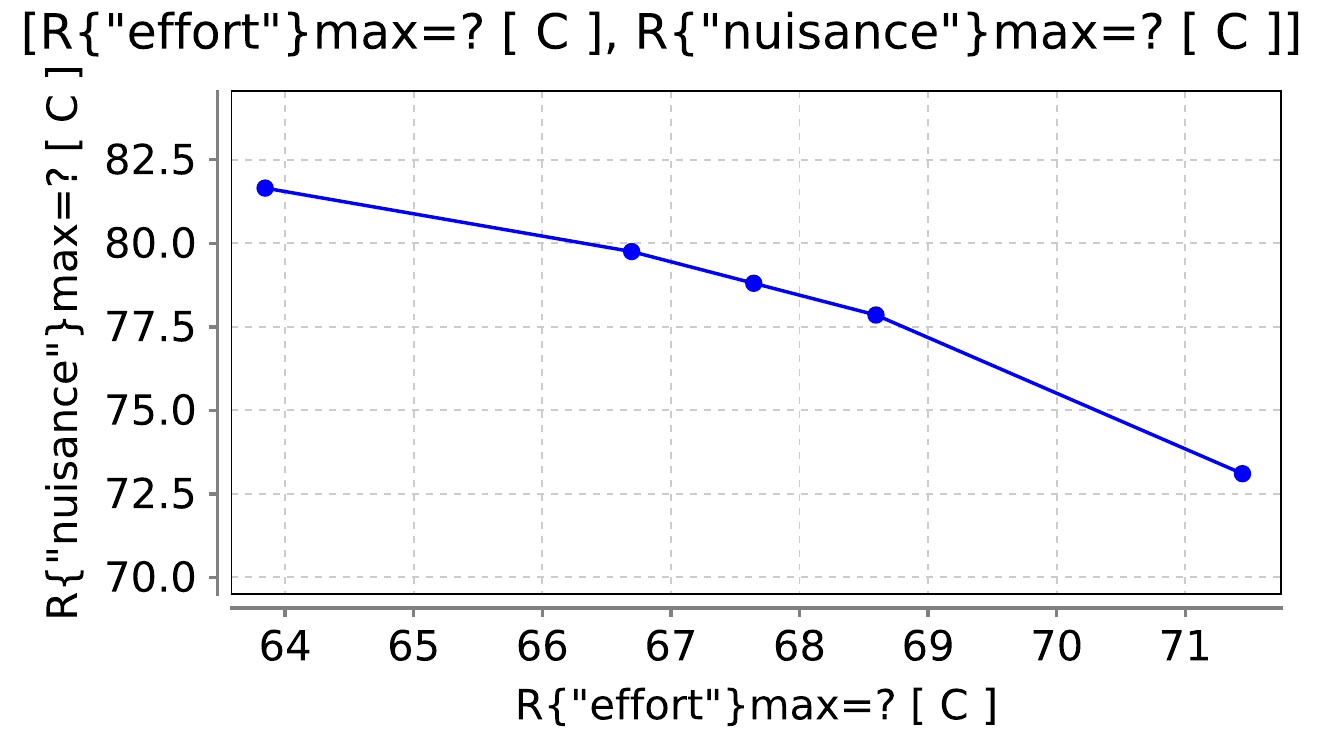}
  \caption{Pareto curve with five policies for \Cref{eq:opt_c} for
    model 7}
  \vspace{-1em}
\end{figure}

For demonstration of \textsc{Yap}\xspace's capabilities, the incident $\rfct[RT]$ and
the accident $\rfct[RC]$ are included in the risk model without
handler commands.  However, these factors add further constraints on
$R(F)$ to be dealt with by the \ac{ASC}.  Hence, $mr$ stays at 15
actions and $c$ rises to $15$ constraints.  In model 7~(last line of
\Cref{tab:exp:results}), the $\Xi$-fraction of $S$~(12079 states) and
$R(F)$~(122 risk states) differ by two orders of magnitude.  We
believe, such an abstraction underpins the potential usefulness of the
proposed risk model in such applications.

For \textbf{RQ3}, we can at the current stage of this project only
provide a ballpark figure for the \emph{detection and handling
  overheads}.  Let $t\colon \alpha_{\mathcal{P}}\to\mathbb{R}$ be the
processing time required for an action, e.g.\xspace for the calculation of
the detection of $\rfct[HC]$ in $\acend{\rfct[HC]}$.  If implemented
as part of a sequential cell controller, the \ac{ASC} requires a time
slot of length $\Sigma_{\rfct\in F} t(\acend{\rfct})$ in each control
cycle.  If monitored simultaneously in dedicated \ac{ASC} hardware,
the slowest detection rate for $F$ is
$1 / \max_{\rfct\in F} t(\acend{\rfct})$.  The overhead for handling
$\rfct$ can be estimated from \Cref{fig:riskfactor} and may range from
$t(\acmit{HC})$ to
$\Sigma_{k\in\{\mathit{sm,a,sf}\}}\big(t(\acmit[k]{\rfct})+t(\acmit[r,k]{\rfct})\big)$.

\subsection{Discussion}
\label{sec:discussion}

\subsubsection*{Relative Safety of a Policy}

To simplify game-theoretic reasoning about $\mathcal{M}$, we reduce
non-deterministic choice for the environment~(i.e.,\xspace operator,
robot, welder).  The more deterministic such choice, the
closer the gap between policy space $\Pi_{\mathcal{M}}$ and 
\ac{ASC} design space.  Any decisions left to the environment will
make a verified policy $\pi$ \emph{safe} relative to $\pi$'s
environmental decisions.  These decisions form the \emph{assumption}
of the \ac{ASC}'s \emph{safety guarantee}.
Occupational health and safety assumes trained operators not to act
maliciously, suggesting ``friendly environments'' with realistic human
errors.  To increase priority of the \ac{ASC}, we can express such an
assumption, e.g.\xspace by minimising \emph{risk} and maximising \emph{pot}.

\subsubsection*{Sensing Assumptions} 
\label{sec:requ-sens-syst}

In our example, the \ac{ASC} relies on the detection of an
\emph{operator}~(e.g.\xspace extremities, body) and a \emph{robot}~(e.g.\xspace arm,
effector) entering a location, the \emph{cell state}~(e.g.\xspace grabber
occupied, workbench support filled), and the \emph{workpiece
  location}~(e.g.\xspace in grabber, in support).  For $\mathcal{M}$, we
assume the tracking system~(i.e.,\xspace range finder and light barrier in the
industrial setting, Kinect %
in the lab replica) to map the \emph{location} of the operator and
robot to the areas ``at table'', ``at workbench'', ``in cell'', and
``at welding spot''.
In \Cref{fig:setting-actual1}, the range finder signals ``at welding
spot'' if the closest detected object is nearer than the close range,
and ``in cell'' if the closest object is nearer than the wide range.
Tracking extensions, not discussed here, could include object
silhouettes and minimum distances, operator intent, or joint velocities
and forces.

\subsubsection*{Sensor Faults}
\ac{pGCL} requires much care with the modelling of real-time behaviour,
particularly, when actions from several concurrent modules
are enabled.
To model real-time \ac{ASC} behaviour, we synchronise operator actions
with sensor events and force the priority of \ac{ASC} reactions in
$\pi^{\star}$ by maximising the risk reduction
potential~(cf.\xspace$\mathit{pot}$ in \Cref{tab:properties}).  While
\emph{synchronisation} restricts global variable use increasing
$\mathcal{M}$'s state space, we found it to be the best solution.

\subsubsection*{Model Debugging and Tool Restrictions}

To reduce the state space, we strongly discretise \emph{location}.
To simplify debugging, we use probabilistic choice in synchronous
updates only in one of the participating commands.
To support synchronisation with complex updates, we avoid global
variables.

State rewards would allow a natural modelling of, e.g.\xspace risk
exposure. In \textsc{PRISM}\xspace~4.5, one needs to use action rewards for
multi-objective queries of \acp{MDP}.
Risk gradient matrices help to overcome a minor restriction in
\textsc{PRISM}\xspace's definition of action rewards.\footnote{Currently, rewards
  cannot be associated with particular updates, i.e.,\xspace with incoming
  transitions rather than only states.}  Alternatively, we could have
introduced extra states, however, at the cost of increasing
$\mathcal{M}$'s state space, undesirable for synthesis.
Rewards require the elimination of non-zero end components~(i.e.,\xspace
deadlocks or components with cycles that allow infinite paths
and, hence, infinite reward accumulation).  \textsc{PRISM}\xspace provides facilities
to identify such components, however, their elimination is non-trivial
and laborious in large models and can require intricate model revisions.

\section{Conclusion}
\label{sec:conclusion}

We introduced a tool-supported method for the correct-by-construction synthesis of
\aclp{ASC} from \acl{MDP} models of \acl{HRC} settings.  These
controllers implement regulatory safety goals for such settings.  We
describe steps for streamlining the modelling of \acp{MDP}.  Our
method draws support from two tools, \textsc{Yap}\xspace for structured risk
modelling and \ac{MDP} generation and \textsc{PRISM}\xspace for probabilistic model
checking and \ac{MDP} policy synthesis.
We show that our approach can be used to incrementally build up
multi-hazard models including alternative mitigation and resumption
options.  
Hence, our approach improves the state of the art of
\ac{ASC} synthesis for \ac{HRC} settings, particularly when dealing
with multiple risks, mitigation options, and safety modes.
The verification results obtained by using our
method can form evidence in an \ac{ASC} \emph{assurance
  case}~\cite{Gleirscher2019-EvolutionFormalModel}.

\subsubsection*{Future Work}

Our approach limits the inference of high effectiveness of an \ac{ASC}
from high conditional accident freedom of the associated policy.  Our
setting can require the assessment of how much the decisions of the
\ac{ASC} and the environment contribute to the accident freedom.  We
plan to explore game-theoretic settings to remove this limitation.

The evaluation of the verified controller in the manufacturing
cell~(e.g.\xspace overhead in resource usage, influence on nominal operation)
is out of scope of this paper.  Such an evaluation requires the
translation of the controller into an executable form.  Our next steps
will be the conversion of the synthesised \ac{DTMC} into a program for
the digital twin simulator and the replica of the cell.  Note that
this translation has to be verified to match the executable form with
the verified properties.  Additionally, we plan to derive tests for
this program from the facilities provided by the simulator.

For optimal synthesis, the proposed method uses parameters such as
upper risk and severity bounds in constraints.  We plan
to introduce parameters for the probabilities into the \ac{MDP},
supported by tools such as
\textsc{evoChecker}\xspace~\cite{Gerasimou2018-Synthesisprobabilisticmodels}, and to
use parametric risk gradient matrices by extending \textsc{Yap}\xspace.
We intend to explore the use of \textsc{evoChecker}\xspace to avoid the split of
the verification procedure into two
stages~(cf.\xspace\Cref{sec:verif-mdp,sec:verif-dtmc}).
We also like to explore online policy synthesis to allow more variety
in environmental decisions~(e.g.\xspace malicious operators).  This
corresponds to weakening the assumptions under which the \ac{ASC} can
guarantee safety.

Unable to collect data~(cf.\xspace\Cref{sec:uncertainties}) from an
industrial application, we had to make best guesses of
probabilities.  However, the frequency of undesired intrusion of
operators into the safeguarded area and accident likelihood can be
transferred into our example.
This example can be extended by randomised control decisions with
fixed probabilities~(e.g.\xspace workload), by adding uncertain action
outcomes~(e.g.\xspace welding errors), and by time-dependent randomised
choice of mitigation options.  To use time in guarded commands, we
want to explore clock-based models as far as synthesis capabilities
allow this, rather than only using reward structures.

\subsubsection*{Acknowledgements}

This research was funded by the Assuring Autonomy International
Programme grant CSI:Cobot.
We are grateful for many insights into manufacturing robot control
from our project partners at the University of Sheffield and our
industrial collaborator.  We also thank David Parker for his
advice in the use of \textsc{PRISM}\xspace's policy synthesis facility.

\bibliographystyle{IEEEtran}
\bibliography{}
\end{document}